\documentclass[lettersize,journal]{IEEEtran}
\usepackage{amsmath,amsfonts}
\usepackage{algorithmic}
\usepackage[ruled]{algorithm2e}
\usepackage{array}
\usepackage[caption=false,font=scriptsize,labelfont=sf,textfont=sf]{subfig}
\usepackage{textcomp}
\usepackage{stfloats}
\usepackage{url}
\usepackage{verbatim}
\usepackage{graphicx}
\usepackage{cite}
\usepackage{booktabs}
\usepackage{multirow}
\usepackage{makecell}
\usepackage{hyperref}
\DeclareSubrefFormat{myparens}{#1(#2)}
\bstctlcite{IEEEexample:BSTcontrol}

\begin{document}

\title{WaterScenes: A Multi-Task 4D Radar-Camera Fusion Dataset and Benchmarks for Autonomous Driving on Water Surfaces}

\author{
Shanliang Yao\textsuperscript{1,*}, Runwei Guan\textsuperscript{1,*}, Zhaodong Wu\textsuperscript{2}, Yi Ni\textsuperscript{2}, Zile Huang\textsuperscript{2}, Ryan Wen Liu\textsuperscript{3}, Yong Yue\textsuperscript{2}, \\ Weiping Ding\textsuperscript{4}, \IEEEmembership{Senior Member,~IEEE}, Eng Gee Lim\textsuperscript{2}, \IEEEmembership{Senior Member,~IEEE}, Hyungjoon Seo\textsuperscript{1}, Ka Lok Man\textsuperscript{2}, Jieming Ma\textsuperscript{2}, Xiaohui Zhu\textsuperscript{2,$\dagger$}, Yutao Yue\textsuperscript{5,$\dagger$}

\thanks{$^{\text{1}}$ Shanliang Yao, Runwei Guan and Hyungjoon Seo are with Faculty of Science and Engineering, University of Liverpool, Liverpool, UK. (email: \{shanliang.yao, runwei.guan, hyungjoon.seo\}@liverpool.ac.uk).}
\thanks{$^{\text{2}}$ Zhaodong Wu, Yi Ni, Zile Huang, Yong Yue, Eng Gee Lim, Ka Lok Man, Jieming Ma and Xiaohui Zhu are with School of Advanced Technology, Xi'an Jiaotong-Liverpool University, Suzhou, China. (email: \{zhaodong.wu20, yi.ni21, zile.huang21\}@student.xjtlu.edu.cn. \{yong.yue, enggee.lim, ka.man, jieming.ma, xiaohui.zhu\}@xjtlu.edu.cn).}
\thanks{$^{\text{3}}$ Ryan Wen Liu is with School of Navigation, Wuhan University of Technology, Wuhan 430063, China, and also with the State Key Laboratory of Maritime Technology and Safety, Wuhan 430063, China (email: wenliu@whut.edu.cn).}
\thanks{$^{\text{4}}$ Weiping Ding is with School of Information Science and Technology, Nantong University, Nantong 226019, China. (email: dwp9988@163.com).}
\thanks{$^{\text{5}}$ Yutao Yue is with Thrust of Artificial Intelligence and Thrust of Intelligent Transportation, The Hong Kong University of Science and Technology (Guangzhou), Guangzhou 511400, China. (email: yutaoyue@hkust-gz.edu.cn).}
\thanks{$^{\text{*}}$ Equal contribution}
\thanks{$^{\dagger}$ Corresponding author: xiaohui.zhu@xjtlu.edu.cn, yutaoyue@hkust-gz.edu.cn}
}

\maketitle

\begin{abstract}
Autonomous driving on water surfaces plays an essential role in executing hazardous and time-consuming missions, such as maritime surveillance, survivor rescue, environmental monitoring, hydrography mapping and waste cleaning. 
This work presents WaterScenes, the first multi-task 4D radar-camera fusion dataset for autonomous driving on water surfaces. Equipped with a 4D radar and a monocular camera, our Unmanned Surface Vehicle (USV) proffers all-weather solutions for discerning object-related information, including color, shape, texture, range, velocity, azimuth, and elevation. Focusing on typical static and dynamic objects on water surfaces, we label the camera images and radar point clouds at pixel-level and point-level, respectively. In addition to basic perception tasks, such as object detection, instance segmentation and semantic segmentation, we also provide annotations for free-space segmentation and waterline segmentation.
Leveraging the multi-task and multi-modal data, we conduct benchmark experiments on the uni-modality of radar and camera, as well as the fused modalities. Experimental results demonstrate that 4D radar-camera fusion can considerably improve the accuracy and robustness of perception on water surfaces, especially in adverse lighting and weather conditions.
WaterScenes dataset is public on \url{https://waterscenes.github.io}.
\end{abstract}


\section{Introduction}

\IEEEPARstart{A}{utonomous} driving techniques are developing rapidly in recent years, achieving safer, more efficient, and more sustainable transportation across roads, skies, and water surfaces \cite{feng2020deep, wu2021deep, bovcon2021mods}. Different scenarios offer unique prospects and challenges for autonomous driving vehicles. Unmanned Surface Vehicles (USVs) that navigate on water surfaces offer a versatile and cost-effective solution for various tasks, including coastal surveillance, environmental monitoring, river modeling, underwater detection, river rescue, and waste cleaning \cite{bovcon2019mastr1325, cheng2021robust, lin2022maritime, wang2023deep}.

\begin{figure}[!t]
\begin{center}
\includegraphics[width=1\linewidth]{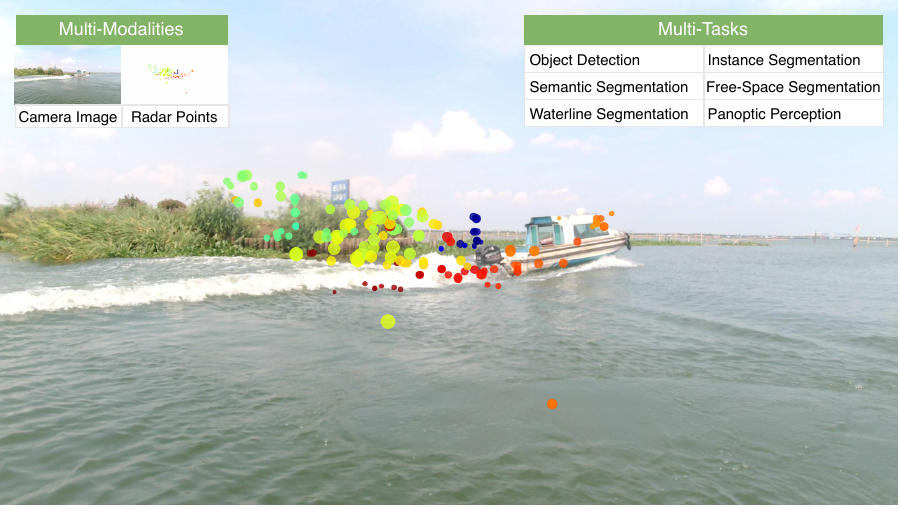}
\end{center}
\vspace{-4mm}
\caption{Example scenario from our WaterScenes dataset. For each radar point on the image, the color denotes the range, and the size represents reflected power from the target.}
\label{fig:cover}
\vspace{-4mm}
\end{figure}

Compared to autonomous driving on road surfaces, perception challenges encountered on water surfaces are more daunting and unpredictable. 
Wind and waves significantly influence the stability of USVs, making it challenging for them to maintain desired heading and trajectory. 
The vibrations produced by USVs have a deleterious effect on sensor output, resulting in blurred transitions from the water to the sky or even object missing in the field of view \cite{wang2023model, he2024active, fang2024dynamic}.
Cameras may be disturbed by water splashes during navigation or water vapor formed due to temperature differences, leading to blurred or obscured images \cite{wang2022review, wang2024aodemar}. 
To further complicate matters, floating debris (e.g., fallen leaves, water plants) along with the rippling caused by waterdrops on water surfaces are distractions to objects of interest.
Mirror-like reflections of water surfaces are challenging to discern between virtual and actual objects.
Adverse lighting and weather conditions significantly impact visibility, further diminishing the clarity of the images \cite{liu2023aioenet, qu2024double, wang2024mdd}. 
These manifold factors present a series of challenges to the camera sensor, making it difficult to detect and track objects in their surroundings.
Although LiDARs can assist in detection accuracy, they are also susceptible to adverse weather conditions \cite{xie2022ammf, li2022modality}. 
Moreover, LiDARs are limited by high waves and water reflectivity when applied to water environments \cite{liu2016unmanned, cheng2021we}.







Unlike camera and LiDAR sensors, radar sensors emit radio waves that bounce off objects and return to the sensor, providing information about the object's range, velocity, and azimuth angle. 
The ability of radar waves to penetrate severe weather conditions with minimal attenuation enables radar sensors to detect objects through rain, fog, and snow \cite{ouaknine2021carrada, bijelic2020seeing, yao2023radarperception}. 
Moreover, the longer wavelength of radar signals makes them less susceptible to interference from adverse lighting conditions, including strong sunlight and darkness \cite{bilik2022comparative}. 
Furthermore, radar sensors can detect objects at long distances and even obstacles behind walls, providing the vehicle with early warning of potential obstacles or hazards \cite{scheiner2020seeing}. 
All these advantages make radar sensors a reliable and robust component in autonomous driving vehicles, equally suitable for overcoming challenges on water surfaces.
Nevertheless, conventional radars possess low resolution and lack semantic information about the detected objects \cite{wang2021rodnet, liu2022deep}. When applied on water surfaces, they produce weak echoes from non-metallic targets, along with clutter returned from the water environments \cite{cheng2021flow, cheng2021robust}.

Therefore, a multi-modal sensor fusion approach that combines the strengths of radar and camera sensors is a potential solution to overcome the challenges and provide a comprehensive understanding of water surface perception.
Numerous studies have demonstrated that multi-sensor fusion can overcome the shortcomings of each individual sensor, improving the overall scene understanding for intelligent transportation systems \cite{zhang2023cmx, berrio2021camera, guo2023asynchronous}. Radar-camera fusion, a typical representative in multi-sensor fusion, has also received considerable attention, demonstrating improved accuracy and robustness of models for autonomous driving vehicles on roads \cite{chadwick2019distant, nabati2021centerfusion, kim2023craft, yao2023radar}.
However, few works focus on radar-camera fusion on water surfaces, primarily due to the lack of available multi-modal datasets. 
To the best of our knowledge, FloW \cite{cheng2021flow} is the only dataset that contains both radar and camera data for USVs. As there is only one category named ``bottle" in FloW dataset, it is unsuitable for the complex water surface environment in real scenarios.

In recent years, 4D radar has shown its advantages in denser radar point clouds and higher angle resolution, providing richer information about the target. Thus, it is a potential perception sensor on USVs, tackling the unique challenges on water surfaces, such as surface reflections, adverse lighting and weather conditions.
An increasing number of 4D radar-camera fusion datasets (e.g., Astyx \cite{meyer2019automotive}, K-Radar \cite{paek2022k}, VoD \cite{palffy2022multi} and TJ4DRadSet \cite{zheng2022tj4dradset}) have emerged for autonomous driving on roads and proved to be effective in improving the accuracy of detection \cite{zheng2023rcfusion, xiong2023lxl}. 
However, there is no public 4D radar dataset for water surfaces till now, let alone a fused 4D radar and camera dataset.
As shown in Fig. \ref{fig:cover}, our proposed dataset fills this gap with the following contributions:

\begin{itemize}
\item We present WaterScenes, the first multi-task 4D radar-camera fusion dataset on water surfaces, which offers data from multiple sensors, including a 4D radar, monocular camera, GPS, and IMU. It can be applied in six perception tasks, including object detection, instance segmentation, semantic segmentation, free-space segmentation, waterline segmentation, and panoptic perception.
\item Our dataset covers diverse time conditions (daytime, nightfall, night), lighting conditions (normal, dim, strong), weather conditions (sunny, overcast, rainy, snowy) and waterway conditions (river, lake, canal, moat). An information list is also offered for retrieving specific data for experiments under different conditions.
\item We provide 2D box-level and pixel-level annotations for camera images, and 3D point-level annotations for radar point clouds. We also offer a toolkit\footnote{\url{https://github.com/WaterScenes/WaterScenes}} for WaterScenes that includes pre-processing, labeling, projection and visualization, assisting researchers in processing and analyzing our dataset.
\item We build corresponding benchmarks and evaluate popular algorithms for object detection, point cloud segmentation, image segmentation, and panoptic perception. Experiments demonstrate the advantages of radar perception on water surfaces, particularly in adverse lighting and weather conditions.
\end{itemize}

The rest of our study is organized as follows: 
Section \ref{sec:Related Datasets} reviews the related datasets on water surfaces, highlighting the significance of our WaterScenes.
Section \ref{sec:WaterScenes} offers detailed insights into the proposed dataset, including USV setup, data collection, data processing, and dataset analysis.
Section \ref{sec:Benchmarks} and Section \ref{sec:Discussions} present benchmark experiments to evaluate the dataset, along with discussions on challenges and potential research directions.
Lastly, in Section \ref{sec:Conclusion}, we summarize our study and provide an outlook for future works.

\section{Related Datasets}
\label{sec:Related Datasets}

\begin{table*}[!h]
\caption{Overview of public datasets on water surfaces. 
($^{\dagger}$) denotes the number of classes in the detection task. 
(-) indicates that no information is provided in the dataset.
OD: Object Detection, LS: waterLine Segmentation, OT: Object Tracking, SS: Semantic Segmentation, FS: Free-Space Segmentation, PS: Panoptic Segmentation, IS: Instance Segmentation, PP: Panoptic Perception.}
\setlength\tabcolsep{5pt} 
\center
\footnotesize
\begin{tabular*}{\linewidth}{p{1.9cm}<{}p{0.7cm}<{\centering}p{1cm}<{\centering}p{0.8cm}<{\centering}p{1.5cm}<{\centering}p{1.6cm}<{\centering}p{2.2cm}<{\centering}p{1.2cm}<{\centering}p{1.2cm}<{\centering}p{1.1cm}<{\centering}p{1.1cm}<{\centering}}
\toprule

\bf{Name} & \bf{Year} & \bf{Camera} & \bf{Radar} & \bf{GPS, IMU} & \bf{Tasks} & \bf{Annotations} & \bf{Classes $^{\dagger}$} & \bf{Annotated Frames} & \bf{Adverse Lighting} & \bf{Adverse Weather}\\\midrule
MODD \cite{kristan2015fast} & 2015 & Mono & - & - & OD, LS & 2D Box, 2D Line & 2 & 4,454 & $\checkmark$ & -  \\
MODD2 \cite{bovcon2018stereo} & 2018 & Stereo & - & GPS, IMU & OD, LS & 2D Box, 2D Line & 2 & 11,675 & $\checkmark$ &  $\checkmark$\\
SMD \cite{moosbauer2019benchmark} & 2019 & Mono & - & - & OD, OT & 2D Box & 10 & 31,653 & $\checkmark$ & - \\
MaSTr1325 \cite{bovcon2019mastr1325} & 2019 & Mono & -  & IMU & SS & 2D Pixel & 4 & 1,325 & $\checkmark$ & $\checkmark$ \\
MODS \cite{bovcon2021mods} & 2021 & Stereo & - & IMU & OD, SS & 2D Box, 2D Line & 3 & 24,090 & $\checkmark$ &  $\checkmark$\\
MID \cite{liu2021efficient} & 2021 & Mono & - & - & OD & 2D Box & 2 & 2,655 & $\checkmark$ &  $\checkmark$ \\
USVInland \cite{cheng2021we} & 2021 & Stereo & - & GPS, IMU & SS, FS & 2D Line & 1 & 700 & $\checkmark$ & $\checkmark$\\
FloW \cite{cheng2021flow} & 2021 & Mono & 3D & - & OD & 2D Box & 1 & 2,000 & $\checkmark$ & -\\
LaRS \cite{vzust2023lars} & 2023 & Mono & - & - & SS, PS & 2D Line & 11 & 4,006 & - & - \\
MVDD13 \cite{wang2024marine} & 2024 & Mono & - & - & OD & 2D Box & 13 & 35,474 & $\checkmark$ & $\checkmark$ \\
\midrule
\textbf{WaterScenes} (Ours) & 2023 & Mono & 4D & GPS, IMU & OD, IS, SS, FS, LS, PP & 2D Box, 2D Pixel, 2D Line, 3D Point & 7 & 54,120 & $\checkmark$ & $\checkmark$  \\
\bottomrule
\end{tabular*}
\label{tab:related-datasets}
\end{table*}

Table \ref{tab:related-datasets} gives an overview of public datasets related to water surfaces.
MODD \cite{kristan2015fast} dataset specifically focuses on obstacle detection in marine environments. It contains 12 marine video sequences, each manually labeled with water edges and obstacles. The specific obstacle classification does not refine the objects in each category, but only classifies them into large and small obstacles. Objects that straddle the water edge are marked as large obstacles, while those entirely located below the water edge are marked as small obstacles.
MODD2 \cite{bovcon2018stereo}, an extended version of MODD, provides synchronized IMU data to assist obstacle detection. Additionally, this dataset includes stereo images, which can be used for stereo verification to further enhance the detection performance.
SMD \cite{moosbauer2019benchmark} contains more specific obstacle categories, including ferry, ship, vessel, speed boat, and sail boat, acquired from both shore and boat. Besides, some data are captured from a near-infrared camera, which can provide image data in low light or even dark conditions.
MaSTr1325 \cite{bovcon2019mastr1325} is a marine semantic segmentation dataset, consisting of 1,325 samples and four pixel-level categories, namely obstacle, water, sky and ignore region.
Moreover, MODS \cite{bovcon2021mods} dataset provides annotations for both detection and segmentation tasks. In this dataset, dynamic obstacles (vessel, person and other) are annotated with bounding boxes, while static obstacles (shore and pier) are annotated by water-obstacle boundaries. 
MID \cite{liu2021efficient} serves as a complementary dataset to the MODD \cite{kristan2015fast} by capturing data in different severe weather conditions that coastal USVs may encounter.
MVDD13 \cite{wang2024marine} dataset contains 13 categories, covering various types of marine vessels in both military and civilian fields. Realistic situations such as class proportions, image diversity, sample independence, and background clutter are considered in MVDD13, thus providing in-depth information for training and testing robust detectors.

The aforementioned datasets are tailored toward the marine environment, which predominantly features vast expanses of water. Conversely, inland rivers are characterized by their narrow and complex shapes, as well as diverse objects present on their surfaces.
Introducing a dataset specifically geared towards inland USVs, USVInland \cite{cheng2021we} dataset serves as a resource for multiple tasks, including SLAM, stereo matching, and water segmentation. Unlike prior datasets such as MODD \cite{kristan2015fast} and MODD2 \cite{bovcon2018stereo}, which solely traced the periphery of the water, USVInland provides comprehensive annotation of the entire water area via polygons. 
LaRS \cite{vzust2023lars} is a large maritime panoptic obstacle segmentation dataset, capturing data from lakes, rivers and seas. Its excellence lies in the diversity of recording locations, scene types, obstacle categories, and acquisition conditions.

To draw attention to floating debris cleaning in inland waterways, FloW \cite{cheng2021flow} dataset is proposed for floating waste detection using both camera and radar sensors. The benchmark in this dataset demonstrated the effectiveness of radar sensors in detecting small objects and their potential for application on water surfaces. However, this dataset has only one category (bottle), and the detection range for the radar sensor is limited to 14.5 meters.

\section{WaterScenes Dataset}
\label{sec:WaterScenes}

\begin{figure}[htbp]
\begin{center}
\includegraphics[width=1\linewidth]{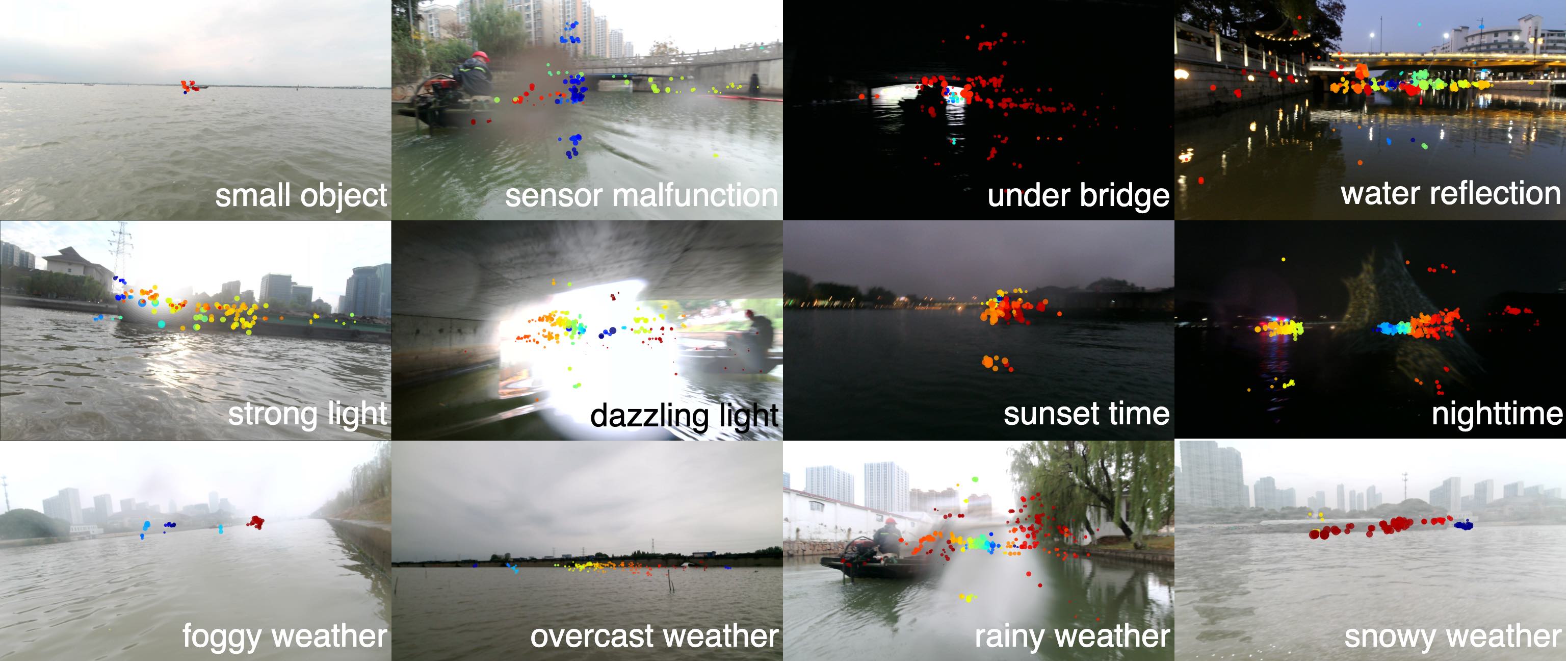}
\end{center}
\vspace{-2mm}
\caption{Samples in WaterScenes. Radar points are projected onto the image plane as colored dots.}
\label{fig:samples}
\end{figure}


As can be intuitively seen from Fig. \ref{fig:samples}, our WaterScenes provides multi-modal and multi-task data for autonomous driving on various water surface scenarios. 
Information about the WaterScenes is summarized in Table \ref{tab:related-datasets}, including equipped sensors, perception tasks, annotation types and collection conditions. 
In this section, we present the process of creating this dataset and provide a statistical analysis of its contents.

\subsection{USV Setup}

\begin{figure}[htbp]
\begin{center}
\includegraphics[width=1\linewidth]{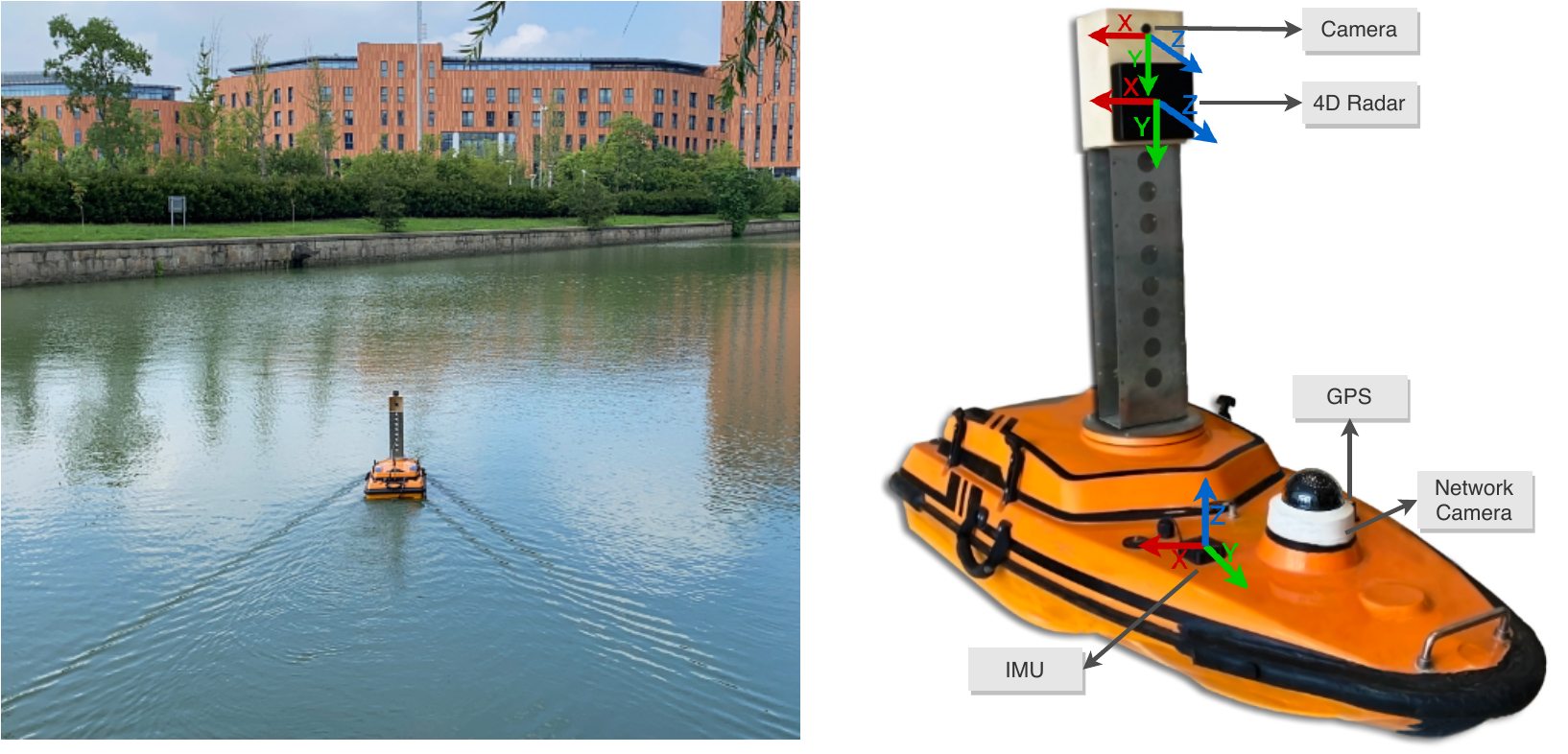}
\end{center}
\vspace{-2mm}
\caption{Sensor suite for our USV and coordinate system of each sensor. }
\label{fig:usv}
\end{figure}

Our USV for data collection is equipped with various sensors, including a 4D radar for capturing radar point clouds, a monocular camera for gathering image information, a network camera for 360-degree observation, a GPS for geographical location information, and an IMU for tracking posture and motion information. 
The arrangement of these sensors on the USV is illustrated in Fig. \ref{fig:usv}, with each sensor's origin and direction denoted by different colors in the coordinate systems. 
The detailed specifications of each sensor mounted on our USV are outlined in Table \ref{tab:sensors}.

\begin{table}[hbt]
\caption{Specifications of sensors equipped on our USV.}
\footnotesize
\begin{tabular*}{\linewidth}{p{0.8cm}  p{7.2cm}}
\toprule
\bf{Sensor} & \bf{Details} \\\midrule
Radar & Oculii EAGLE $77\text{GHz}$ Point Cloud Radar, Medium Range Mode: $200\mathrm{~m}$ detection range, $0.43\mathrm{~m}$ range resolution, $0.27\mathrm{~m}$/$\mathrm{s}$ velocity resolution, $<1^\circ$ azimuth/elevation angle resolution, $110^\circ$ HFOV, $45^\circ$ VFOV, $15\text{Hz}$ capture frequency\\
\vspace{-1mm} Camera & \vspace{-1mm} SONY IMX317 CMOS sensor, RGB color, $1920 \times 1080$ resolution, $100^\circ$ HFOV, $60^\circ$ VFOV, $30\text{Hz}$ capture frequency\\
\vspace{-1mm} GPS & \vspace{-1mm} latitude, longitude and altitude coordinates, $<2.5\mathrm{~m}$ position accuracy, $<0.1\mathrm{~m}$/$\mathrm{s}$ velocity accuracy, $10\text{Hz}$ update rate\\
\vspace{-1mm} IMU & \vspace{-1mm} 10-axis inertial navigation ARHS (3-axis gyroscope, 3-axis accelerometer, 3-axis magnetometer and a barometer), $0.5^\circ$ heading accuracy, $0.1^\circ$ roll/pitch accuracy, $50\text{Hz}$ update rate\\
\bottomrule
\end{tabular*}

\label{tab:sensors}
\end{table}



\begin{table*}[!h]
\caption{Dataset statistics. Number of annotated frames (top), number of objects (middle), and percentage of objects belonging to each class compared to the total number of objects (bottom). ($^{\dagger}$) Free-Space class is included in instance segmentation, semantic segmentation and panoptic perception tasks. ($^{\dagger\dagger}$) Waterline annotations are in waterline segmentation and panoptic perception tasks.}
\setlength\tabcolsep{5pt} 
\center
\footnotesize
\begin{tabular*}{\linewidth}{p{1.3cm}<{}p{0.9cm}<{\centering}|p{0.9cm}<{\centering}p{0.9cm}<{\centering}p{0.9cm}<{\centering}p{0.9cm}<{\centering}p{0.9cm}<{\centering}p{0.9cm}<{\centering}p{0.9cm}<{\centering}|p{1.4cm}<{\centering}p{1.3cm}<{\centering}|p{1.1cm}<{\centering}p{1.1cm}<{\centering}}
\toprule

 & \bf{Total} & \bf{Pier} & \bf{Buoy} & \bf{Sailor} & \bf{Ship} & \bf{Boat} & \bf{Vessel} & \bf{Kayak} & \bf{Free-Space} $^{\dagger}$ & \bf{Waterline} $^{\dagger\dagger}$ & \bf{Adverse Lighting} & \bf{Adverse Weather} \\\midrule
Frames & 54,120 & 25,787 & 3,769 & 3,613 & 19,776 & 9,106 & 9,362 & 366 & 54,057 & 53,926 & 5,604 & 10,729  \\
Objects Percentage & 202,807 & 121,827 (60.07\%) & 16,538 (8.15\%) & 8,036 (3.96\%) & 34,121 (16.82\%) & 10,819 (5.33\%) & 11,092 (5.47\%) & 374 (0.18\%) & 54,057 & 159,901 & 30,517 (15.05\%) & 46,784 (23.07\%)    \\
\bottomrule
\end{tabular*}
\label{tab:waterscenes-statistics}
\end{table*}

\subsection{Data Collection}
Our dataset is collected from June to December 2022 in Suzhou, China. 
As the objects and surrounding environments vary across different water conditions, we select various waterways for data collection, such as small and large rivers, lakes, canals and moats.
In order to capture high-quality data, we employ two distinct control methods during the data collection process. 
The first method utilizes our custom-designed software to create a precise navigation path, allowing the USV to travel to a specific location while recording data without human intervention. The second method involves remote control, which is used to acquire data for specific objects from multiple viewpoints.
We focus on common objects of interest on water surfaces, including static objects such as piers and buoys, and dynamic objects such as ships, boats, vessels, kayaks, and sailors aboard these surface vehicles.

Meanwhile, to ensure the diversity and comprehensiveness of the dataset, we collect data across different waterways under different time conditions (e.g., daytime, nightfall and night), diverse lighting (e.g., normal, dim and strong) and weather conditions (e.g., sunny, overcast, rainy and snowy).
We also document scenarios of sensor malfunction, including instances where waterdrops adhere to the camera lens, resulting in obscured images, as well as situations where radar connectivity is lost, rendering it impossible to capture point cloud data. These records are significant as they reflect real-world challenges that are likely to arise during autonomous driving.

\subsection{Processing and Annotation}
Following the processing approach from the nuScenes dataset \cite{caesar2020nuscenes}, we extract image keyframes at a rate of 2Hz. The radar, GPS, and IMU data are then synchronized with the image keyframes based on the closest timestamp, with a maximum tolerated time difference of 0.05 seconds \cite{palffy2022multi}.
Each image in the dataset is manually annotated by human annotators and is further validated by domain experts. 
In the object detection task, seven categories (pier, buoy, sailor, ship, boat, vessel and kayak) are enclosed in each image by 2D bounding boxes. 
For the instance segmentation task, an additional category named free-space is labeled using polygonal masks, which indicates drivable areas for USVs.
Annotations for semantic segmentation and free-space segmentation are later generated using the instance segmentation labels. 
To facilitate the waterline segmentation task, we draw lines that mark the boundary between water and land.
In addition to annotating the class for each object, we also label the attributes (such as waterways, time conditions, lighting conditions, and weather conditions) for each frame in an information list, which facilitates the retrieval of specific data and the selection of desired data for experiments.


Annotation process for radar point clouds is extremely complicated and tedious, while annotation precision is essential to model training. 
Each point within the radar point clouds comprises various attributes, including range, Doppler velocity, azimuth angle, elevation angle, and reflected power.
To establish the relationship between radar point clouds and camera images, we convert radar point clouds from Polar coordinates onto the image plane utilizing the extrinsic matrix between the radar sensor and camera sensor as well as the intrinsic matrix of the camera sensor \cite{domhof2021joint}.
With coordinates of radar point clouds corresponding to the image plane, we annotate point clouds within the image bounding box as the same category as the box.
However, it should be noted that while this approach can provide some initial annotations for the radar point clouds, these annotations may not always be accurate due to the nature of radar detection. 
Radar point clouds may not consistently map onto objects and may detect targets behind them \cite{meyer2019deep, stacker2022fusion}. Therefore, annotations are refined by domain experts based on projections derived from front and bird's eye views, along with attributes (reflected power and Doppler velocity) of each point. 
Finally, every point within the radar point clouds is assigned a class label and an instance identification.
Furthermore, we include radar data from three and five consecutive frames in WaterScenes, providing valuable resources for analyzing multi-frame accumulation techniques.

\subsection{Dataset Statistics}

WaterScenes dataset includes 54,120 sets of RGB images, radar point clouds, GPS and IMU data, covering over 200,000 objects. The specific number of frames and objects for each class is shown in Table \ref{tab:waterscenes-statistics}. Additionally, as an essential part of this dataset, images captured under unfavorable lighting and weather conditions are enumerated in the table. All images are in 1920 $\times$ 1080 pixels, containing a diverse range of objects, including piers, buoys, sailors, ships, boats, vessels and kayaks. Among the categories, buoys and piers are noticeable obstacles on water that USVs should avoid while driving, whereas ships, boats, vessels and kayaks represent common watercraft encountered on water surfaces. The term ``sailor" specifically refers to the individuals on these watercraft.

\begin{figure}[h]
\centering
\subfloat[Size distribution]{
\includegraphics[width=0.5\columnwidth]{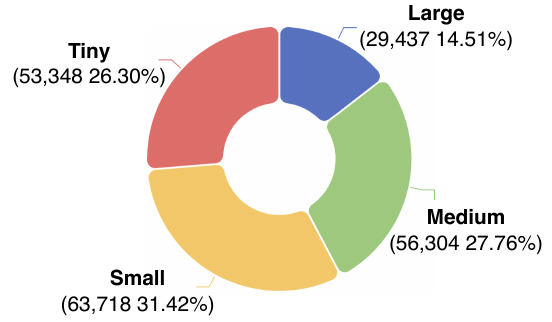}
\label{fig:size}
}
\quad
\hspace{-6mm}
\subfloat[Distance distribution]{
\includegraphics[width=0.45\columnwidth]{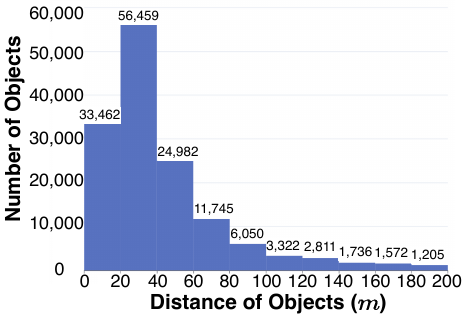}
\label{fig:distance}
}
\caption{Statistics of objects in WaterScenes. (a) Wide range of object size. (b) Wide distribution of object distance.}
\label{fig:statistics}
\end{figure}

We classify objects based on their size as follows: those with an area greater than 192 $\times$ 192 pixels are considered large, those with an area less than 32 $\times$ 32 pixels are deemed tiny, objects with an area between 32$^2$ and 64$^2$ pixels are referred to as small, and those between 64$^2$ and 192$^2$ pixels are classified as medium.
The size distribution depicted in Fig. \subref*{fig:size} reveals a wide range of object sizes, consistent with the diverse sizes of objects typically observed on water surfaces. 
We also analyze the distance distribution using the range attribute in radar point clouds. Fig. \subref*{fig:distance} demonstrates the relationship between the number of objects and distance at intervals of every 20 meters.

\begin{table}[htbp]
\caption{Point cloud attributes for each category.}
\setlength\tabcolsep{4pt}
\center
\footnotesize
\begin{tabular*}{\linewidth}{p{1.8cm}<{}|p{0.6cm}<{\centering}p{0.7cm}<{\centering}p{0.7cm}<{\centering}p{0.7cm}<{\centering}p{0.7cm}<{\centering}p{0.7cm}<{\centering}p{0.7cm}<{\centering}}
\toprule
\bf{Attribute} & \bf{Pier} & \bf{Buoy} & \bf{Sailor} & \bf{Ship} & \bf{Boat} & \bf{Vessel} & \bf{Kayak}  \\\midrule
Points & 8.45 & 14.53 & 4.75 & 81.23 & 38.51 & 80.32 & 6.72\\
Power ($dB$)& 13.68 & 17.88 & 12.15 & 14.40 & 14.14 & 13.52 & 10.12   \\
Velocity ($m/s$) & 0.08 & 0.09 & 0.79 & 1.08 & 0.40 & 2.21 & 0.88\\
\bottomrule
\end{tabular*}
\label{tab:radar-statistics}
\end{table}

Furthermore, we conduct a comprehensive analysis of the radar point clouds by calculating the average values of attributes for each specific class. 
As illustrated in Table \ref{tab:radar-statistics}, the number of points is highly correlated with object size. In particular, ships and vessels, being large objects, have the highest number of points, while sailors and kayaks, being small objects, have few points.
Reflected power is similar for piers, ships, boats and vessels as they are primarily composed of cement. Buoys have higher power values as they are made of metal materials, while kayaks are composed of plastic materials with low power values.
Velocity information is also instrumental in distinguishing between different types of objects. For example, stationary targets such as piers and buoys exhibit minimal velocity, while ships and vessels have relatively higher velocities.
Above all, each attribute represents distinct target characteristics and is crucial for point cloud classification.

\begin{figure*}[h]
\centering
\includegraphics[width=1\linewidth]{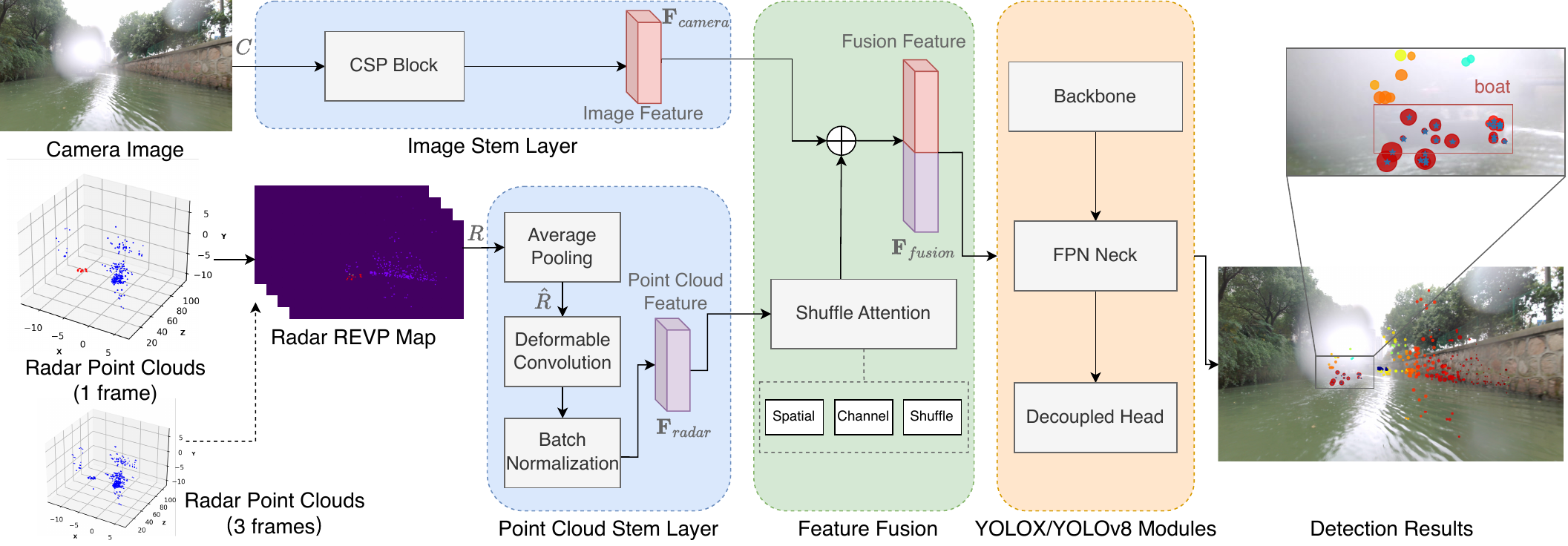}
\caption{Radar-camera fusion network for the detection benchmark on WaterScenes. Camera images and radar point clouds are fed into the stem layers for feature extraction. Subsequently, the extracted features are processed by the attention mechanism and added along the channel dimension before forwarding into YOLOX-M and YOLOv8-M modules. As a result, the fusion-based network successfully detects boats even when cameras are occluded by waterdrops.}
    \label{fig:network}
\end{figure*}

\section{Benchmarks}
\label{sec:Benchmarks}

In this section, our WaterScenes serves as benchmarks for evaluating the performance on multiple tasks on water surfaces, including object detection, radar point cloud segmentation, camera image segmentation and panoptic perception. By analyzing the experimental results, we highlight the value, challenges and potential research directions posed by WaterScenes for further research.

\subsection{Experimental Settings}

After data processing and annotation, we divide the proposed dataset into three parts: a training set, a validation set, and a test set in the ratio of 7:2:1. All experiments are performed on two RTX 3090 GPUs with the training mode of data distributed parallel. All images in WaterScenes are resized to 640 $\times$ 640 pixels during the training phase. Results are evaluated on the test set in WaterScenes with Frames Per Second (FPS) assessed on a single RTX 3090 GPU. 

\textbf{Object Detection.} We select five models for camera-based object detection with diverse paradigms (e.g., two-stage/one-stage, anchor-based/anchor-free, CNN-based/Transformer-based): CenterNet (ResNet-50) \cite{zhou2019objects}, Deformable DETR (ResNet-50)  \cite{zhu2020deformable}, Faster R-CNN (ResNet-50)  \cite{girshick2015fast},  YOLOX-M \cite{ge2021yolox} and YOLOv8-M \cite{yolov8}. We train these models from scratch with an initial learning rate of 1e-2, accompanied by a cosine learning rate scheduler. We set the batch size to 32 and choose Adam as the optimizer with weight decay of 5e-4. We also adopt Exponential Moving Average (EMA) to smooth model weights and mixed precision to speed up the training and reduce the CUDA memory.
 
For fusion-based object detection, we propose a generalized lightweight early fusion method for YOLOX-M and YOLOv8-M without altering their basic architectures. As depicted in Fig. \ref{fig:network}, the detection process incorporates two input modalities: camera RGB images $C \in \mathbb{R}^{3\times H \times W}$ and radar REVP maps $R \in \mathbb{R}^{4\times H \times W}$. Specifically, as described in Algorithm \ref{alg:algorithm}, REVP maps capture the combined features of Range (R), Elevation (E), Velocity (V) and reflected Power (P) of the detected object from the radar point clouds matched to the image frame. 
The coordinate transformation process utilizes an extrinsic matrix accounting for the relative position and orientation of the radar and camera sensors. Subsequently, 3D coordinates in the camera frame are projected onto the 2D image plane using the camera's intrinsic matrix, yielding image plane coordinates $(u, v)$.

In stem layers of the camera input, we follow the default settings of YOLOX \cite{ge2021yolox} and YOLOv8 \cite{yolov8} to conduct the stem step for initial feature downsampling and channel expansion. We then obtain the shallow feature of image $\mathbf{F}_{camera} \in \mathbb{R}^{C \times \frac{H}{2} \times \frac{W}{2}}$, as is shown in Equation \ref{eq:csp}.

\begin{equation}
    \begin{aligned}
  &  \mathbf{F}_{camera} = W_{stem}C,
    \label{eq:csp}
\end{aligned}
\end{equation}
where $W_{stem}$ is the learnable weight of stem layer.

In the case of radar input, we first employ Average Pooling with a window size of $3 \times 3$ and padding value of $1$ to rapidly aggregate sparse neighborhood point clouds $\hat{R}$ (Equation \ref{eq:pool}). Subsequently, we utilize the Deformable Convolution \cite{zhu2020deformable} to extract the irregular radar point cloud features (Equation \ref{eq:radar_stem}). Following this, we apply a Batch Normalization layer, resulting in the radar feature $\mathbf{F}_{radar} \in \mathbb{R}^{C \times \frac{H}{2} \times \frac{W}{2}}$.


\begin{equation}
\begin{aligned}
  &  \hat{R}_{i,j} = \frac{1}{9} \sum_{f=1}^{3} \sum_{g=1}^{3} R_{i+f-1, j+g-1},
  \label{eq:pool}
\end{aligned}
\end{equation}

\begin{equation}
\begin{aligned}
  &  \mathbf{F}_{radar} = \sum\limits_{k=1}^{K} w_k \cdot \hat{R}(p+p_k + \Delta p_k) \cdot \Delta m_k,
    \label{eq:radar_stem}
\end{aligned}
\end{equation}
where $K$ is the convolution kernel of the sampling location. In our experiments, we set $K=9$ as we use a $3 \times 3$ kernel size. $p$ is the pre-specified offset of feature map $\hat{R}$ for $K$ locations. $\Delta p_k$ and $\Delta m_k$ are the learnable offset and modulation scalar for $k$-th location, respectively.

\begin{equation}
\begin{aligned}
  &  \mathbf{F}_{fusion} = \mathbf{F}_{camera} + \alpha \cdot SA(\mathbf{F}_{radar}),
    \label{eq:fusion_feature}
\end{aligned}
\end{equation}
%
%
%
%
%
%
%
%
%
%
%
%
%

\begin{algorithm}
\caption{Radar-Camera Fusion Algorithm}
\label{alg:algorithm}
\begin{algorithmic}[1]

\STATE /* \textbf{Prepare radar REVP maps} */\\
\KwIn{Radar frames (1 frame or 3 frames), number of radar frames $N_r$;}
\KwOut{Radar REVP maps;}
\STATE $Features \gets [Range, Elevation, Velocity, Power]$
\FOR{i $\leftarrow 1$  \TO $N_r$}
\STATE /* Project each radar point onto  camera plane */
\STATE $u, v$ $\gets$ coordinates for radar point on camera plane
\FOR{$channel$ \textnormal{\textbf{in}} $Features$}
\STATE $R_i[channel][u][v]$ $\gets$ radar point channel value
\ENDFOR
\ENDFOR

\STATE  /* \textbf{Set the training stage} */\\
\KwIn{Camera images with annotations, radar REVP maps;}
\KwOut{Radar-camera fusion model;}
\STATE Number of epochs $N_e$ $\leq$ 100;

\FOR{i $\leftarrow 1$ \TO $N_e$}
\STATE /* Feature initialization of camera input $C$ */
\STATE Convolution: $\mathbf{F}_{camera} \leftarrow$ Equation \ref{eq:csp}

\STATE /* Feature initialization of radar input $R$ */

\STATE Average Pooling: $\hat{R} \gets$ Equation \ref{eq:pool}
\STATE Deformable Convolution: $\mathbf{F}_{radar} \gets$ Equation \ref{eq:radar_stem}
\STATE Batch Normalization: $\mathbf{F}_{radar}$ 
\STATE /* Feature fusion upon camera and radar */

\STATE Adaptive Feature Fusion: $\mathbf{F}_{fusion} \gets$ Equation \ref{eq:fusion_feature}

\STATE Run YOLOX/YOLOv8 modules
\ENDFOR

\end{algorithmic}
\end{algorithm}

To mitigate the negative impacts of clutter in radar point clouds while focusing on object features, we apply shuffle attention \cite{zhang2021sa} on $\mathbf{F}_{radar}$, which is a lightweight attention module combining spatial and channel attention. The shuffle operation enhances channel interaction and alleviates over-dependency between inter-layer channels. Moreover, considering that radar plays different roles in various scenarios, assisting the camera modality in some cases and struggling with noise interference in others, we introduce a learnable dynamic weight $\alpha$ to balance the importance of the current sample in the REVP map. 
The outputs from both branches are then element-wise added to generate fused features $\mathbf{F}_{fusion}$, as illustrated in Equation \ref{eq:fusion_feature}. 
After that, we follow the paradigms of YOLOX and YOLOv8 for the backbone, neck and detection head. Overall, the pseudo-code of the proposed radar-camera fusion algorithm is presented in Algorithm \ref{alg:algorithm}.

\textbf{Radar Point Cloud Segmentation.} We select four point cloud processing models with various paradigms, including PointMLP \cite{ma2022rethinking}, Point-NN \cite{zhang2023parameter}, PointNet++ \cite{qi2017pointnet++} and Point Transformer \cite{zhao2021point}. PointMLP \cite{ma2022rethinking}, one of the State-Of-The-Art (SOTA) models in 3D point cloud processing, serves as the primary model for detailed analysis. We train all models from scratch with an initial rate of 5e-4, accompanied by a cosine learning rate scheduler. The batch size is 128 with AdamW as the optimizer and a weight decay of 5e-4. We employ the negative log-likelihood loss with focal \cite{lin2017focal} as the loss function. 

\textbf{Camera Image Segmentation.} We select four classical models with various paradigms: DeepLabv3+ (atrous convolution with ASPP) \cite{chen2018encoder}, HRNet-W32 (multi-scale fusion with high-resolution features) \cite{wang2020deep}, SegNeXt-B (convolution-attention-based) \cite{guo2022segnext}, SegFormer-B1 (self-attention-based) \cite{xie2021segformer} and Mask2Former-R50 (transformer-based all-in-one model) \cite{cheng2022masked} for image semantic segmentation; and another four models with different paradigms: YOLACT (two-stage of localization and segmentation) \cite{bolya2019yolact}, SOLO (one-stage without localization) \cite{wang2020solo}, YOLOv5-M (anchor-based) \cite{yolov5}, YOLOv8-M (anchor-free) \cite{yolov8} and Mask2Former (transformer-based all-in-one model) \cite{cheng2022masked} for image instance segmentation. We train these models from scratch with an initial learning rate of 9e-3, accompanied by a cosine learning rate scheduler. We adopt the dice loss for semantic segmentation and the focal loss for instance segmentation. We set the batch size to 32 and choose SGD as the optimizer with the weight decay of 1e-4 and momentum of 0.937. Moreover, we adopt mixed precision to accelerate the training process and reduce the CUDA memory.  


\begin{table*}[!h]
\caption{Benchmark results of object detection on WaterScenes. In the Modalities column, C denotes the modality from the camera sensor, R denotes the modality from the 4D radar sensor, n-$\textup{frame(s)}$ denotes the accumulation of n-frame radar point clouds. Adverse lighting and weather conditions are evaluated using mAP$_{50}$ metric.}
\center
\footnotesize
\begin{tabular*}{\linewidth}{
p{2.7cm}<{}
p{1.6cm}<{\centering}|
p{0.9cm}<{\centering}
p{0.6cm}<{\centering}
p{0.6cm}<{\centering}|
p{0.55cm}<{\centering}
p{0.55cm}<{\centering}
p{0.55cm}<{\centering}
p{0.55cm}<{\centering}
p{0.55cm}<{\centering}
p{0.55cm}<{\centering}
p{0.65cm}<{\centering}|
p{1cm}<{\centering}
p{0.9cm}<{\centering}}
\toprule
\bf{Model} & \bf{Modalities} & \bf{mAP$_{50\text{-}95}$} & \bf{mAP$_{50}$} & \bf{FPS} &  \textbf{Pier} & \textbf{Buoy} & \textbf{Sailor} & \textbf{Ship} & \textbf{Boat} & \textbf{Vessel} & \textbf{Kayak} & \bf{Adverse lighting} & \bf{Adverse weather}\\\midrule
Faster R-CNN \cite{girshick2015fast} & C & 47.8 & 81.1 & 31.5 & 81.3 & 78.4 & 75.6 & 93.0 & 88.9 & 92.2 & 58.4 & 69.4 & 71.1\\
CenterNet \cite{zhou2019objects} & C & 54.7 & 82.9 & \textbf{117.4} & 83.0 & 80.1 & 79.3 & 92.7 & 89.5 & 93.1 & 62.9 & 72.2 & 73.7\\
Deformable DETR \cite{zhu2020deformable}  & C & 56.5 & 84.0 & 18.2 & 83.9 & 82.2 & 80.2 & 92.9 & 89.4 & 92.7 & 66.8 & 74.5 & 76.2\\
YOLOX-M \cite{ge2021yolox} & C & 57.8 & \textbf{85.1} & 54.7 & \bf{85.1} & 81.1 & 80.5 & 91.4 & 89.5 & 92.1 & \bf{76.1} & \bf{77.4} & 78.9\\
YOLOv8-M \cite{yolov8} & C & \textbf{59.2} & 84.4 & 58.8 & 80.6 & \bf{84.3} & \bf{82.1} & \bf{93.7} & \bf{90.8} & \textbf{95.8} & 62.5 & 74.8 & \bf{79.5}\\
\midrule
YOLOX-M \cite{ge2021yolox} & C + R$_{1\mbox{-}\textup{frame}}$ & 59.5 & 86.1 & 51.2 & 85.5 & 82.2 & 81.3 & 92.9 & 91.3 & 92.5 & 77.1 & 79.8 & 82.5\\
YOLOX-M \cite{ge2021yolox} & C + R$_{3\mbox{-}\textup{frames}}$ & 60.3 & 87.4 & 51.2 & \textbf{87.1} & 84.1 & 86.5 & 93.7 & 91.8 & 91.2 & 77.7 & 81.5 & 83.5\\
YOLOv8-M \cite{yolov8} & C + R$_{1\mbox{-}\textup{frame}}$ & 61.2 & 88.0 & 54.2 & 86.2 & 85.9 & 85.1 & \textbf{94.6} & 91.2 & 95.0 & 77.9 & 80.1 & 82.4\\
YOLOv8-M \cite{yolov8} & C + R$_{3\mbox{-}\textup{frames}}$ & \textbf{62.5} & \textbf{88.8} & 54.2 & 84.5 & \textbf{87.2} & \textbf{87.1} & 94.1 & \textbf{93.2} & \bf{96.3} & \textbf{79.5} & \bf{82.1} & \bf{84.2}\\
\bottomrule
\end{tabular*}
\label{tab:object-detection-baselines}
\end{table*}

\textbf{Panoptic Perception.} In our experiments, the panoptic perception includes tasks of object detection, free-space segmentation and waterline segmentation, covering an all-round perception of water surfaces. We evaluate the performance of panoptic perception on WaterScenes using two camera-based networks (YOLOP \cite{wu2022yolop}, HybridNets \cite{vu2022hybridnets}) and one fusion-based network named Achelous \cite{guan2023achelous}. 
YOLOP and HybridNets comprise one encoder for feature extraction and three decoders to handle the panoptic tasks. Achelous \cite{guan2023achelous} is a lightweight panoptic perception framework dedicated to water surfaces. 
In Achelous, we select MobileViT \cite{mehta2021mobilevit} as the backbone and Ghost Dual-FPN (GDF) as the neck. Besides, we select Radar Convolution \cite{guan2023achelous} to extract radar point cloud features. Furthermore, the homoscedastic-uncertainty-based learning strategy \cite{kendall2018multi} is applied to assist multi-task learning.
In the training stage, the detection head poses challenges in early convergence with an end-to-end strategy. Hence, following the approaches in \cite{wu2022yolop} and \cite{vu2022hybridnets}, we first train the encoder and detection head for 100 epochs. Then, we freeze the encoder and detection head as well as train free-space and waterline segmentation heads for 50 epochs. Finally, the entire network is jointly trained for 50 epochs across all three tasks.

\subsection{Metrics Settings}
This section elaborates on the metrics utilized for evaluating WaterScenes across different tasks. 

\textbf{Object Detection.} We adopt the mean Average Precision (mAP) with an Intersection-over-Union (IoU) threshold of 0.5, denoted as mAP$_{\text{50}}$, and the mAP with an IoU threshold range of 0.5 to 0.95, denoted as mAP$_{\text{50-95}}$. Mathematical formulations of these metrics are presented in Equation \ref{eq:map50} and Equation \ref{eq:map50-95}, respectively, serving as quantitative indicators of a model's effectiveness in detecting objects using bounding boxes.

\begin{equation}
    P = \frac{TP}{TP+FP},
    \label{eq:precision}
\end{equation}

\begin{equation}
    R = \frac{TP}{TP+FN},
    \label{eq:recall}
\end{equation}

\begin{equation}
    AP = \int_{0}^{1} P(r) \, \mathrm{d}r,
    \label{eq:ap}
\end{equation}

\begin{equation}
    \text{mAP}_{50} = \frac{1}{N} \sum_{i=1}^{N} \text{AP}_{50}^i,
    \label{eq:map50}
\end{equation}

\begin{equation}
    \text{mAP}_{\text{50-95}} = \frac{1}{N} \sum_{i=1}^{N} \text{AP}_\text{50-95}^i.
    \label{eq:map50-95}
\end{equation}
$P$ and $R$ correspond to precision and recall as outlined in Equation \ref{eq:precision} and Equation \ref{eq:recall}, respectively. Here, $TP$, $FP$ and $FN$ represent predicted samples of true positive, false positive, and false negative, respectively. 
$AP$ symbolizes the average precision in Equation \ref{eq:ap}, where $P(r)$ denotes the precision on the recall-precision curve and $r$ represents the recall.  
In the equation of \text{mAP}$_{50}$, \text{AP}$_{50}^i$ stands for the AP value of class $i$ targets with an IoU of 50\% and above with ground truth in the predicted bounding boxes. \text{mAP}$_{\text{50-95}}$ denotes the average AP value of class $i$ targets with an IoU ranging from 50\% to 95\% in the prediction box compared to the ground truth.

\textbf{Radar Point Cloud Segmentation.} We adopt Point Accuracy (PA) and mean Intersection-over-Union (mIoU) to evaluate the performances of radar point cloud semantic segmentation, as shown in Equation \ref{eq:pa} and Equation \ref{eq:p_miou}.

\begin{equation}
    \text{PA} = \frac{C}{T},
    \label{eq:pa}
\end{equation}

\begin{equation}
    \text{IoU} = \frac{I}{U},
    \label{eq:iou}
\end{equation}

\begin{equation}
    \text{mIoU} = \frac{1}{N} \sum_{i=1}^{N} \text{IoU}_i,
    \label{eq:p_miou}
\end{equation}
where $C$ represents the number of correctly classified point clouds and $T$ represents the total number of point clouds. 
In the equations for IoU and mIoU, $I$ represents the number of intersection points, $U$ represents the number of union points, and $N$ represents the number of categories.

\textbf{Camera Image Segmentation.} For image semantic segmentation, Overall Accuracy (OA), Mean Pixel Accuracy (MPA) and mIoU are introduced as illustrated in Equation \ref{eq:image_oa}, Equation \ref{eq:image_mpa} and Equation \ref{eq:image_miou}, respectively. For image instance segmentation, mAP$_{50}$ and mAP$_{\text{50-95}}$ in box and mask are employed similarly to the object detection metrics as described in Equation \ref{eq:map50} and Equation \ref{eq:map50-95}.

\begin{equation}
    \text{OA} = \frac{TP + TN}{TP + TN + FP + FN},
    \label{eq:image_oa}
\end{equation}

\begin{equation}
    \text{MPA} = \frac{1}{C} \sum_{i=1}^{C} \frac{TP_i}{TP_i + FP_i},
    \label{eq:image_mpa}
\end{equation}

\begin{align}
    & \text{IoU}_i = \frac{TP_i}{TP_i + FP_i + FN_i}, 
    \label{eq:image_iou}
    \\
    & \text{mIoU} = \frac{1}{C} \sum_{i=1}^{C} \text{IoU}_i,
    \label{eq:image_miou}
\end{align}
where $TP$, $TN$, $FP$ and $FN$ denote predicted samples of true positive, true negative, false positive and false negative, respectively. OA represents the classification accuracy for the whole image and is the proportion of correctly classified pixels among all pixels. MPA is the average pixel accuracy across all classes, which is the proportion of correctly classified pixels to the total number of pixels in that class. Specifically, $C$ represents the number of categories. $TP_i$, $FP_i$ and $FN_i$ are the true positive, false positive and false negative numbers for the $i$-th category, respectively. IoU$_i$ denotes the IoU value for the $i$-th category.

\textbf{Panoptic Perception.} Panoptic perception includes three tasks: object detection, free-space segmentation and waterline segmentation. We adopt mAP$_{50}$ (Equation \ref{eq:map50}) and mAP$_{\text{50-95}}$ (Equation \ref{eq:map50-95}) to evaluate object detection performance. Additionally, we utilize OA (Equation \ref{eq:image_oa}) and mIoU (Equation \ref{eq:image_miou}) to evaluate the free-space segmentation and waterline segmentation tasks.

\subsection{Object Detection}

\textbf{Baseline.} Table \ref{tab:object-detection-baselines} categorizes the object detection baselines into two sections: camera-based detection and fusion-based detection.
For camera-based detection, YOLOv8-M achieves the highest mAP$_{50\text{-}95}$ of 59.2\%, 1.4\% higher than YOLOX-M and 2.7\% higher than Deformable DETR. Besides, it is worth noting that YOLOX-M gets 85.1\% mAP$_{50}$, the highest among all detectors. CenterNet gets the fastest inference speed among all detectors, reaching an impressive 117.4 FPS. 
Furthermore, we evaluate the performance of the models on images captured in challenging lighting and weather conditions. Notably, the accuracy of all models decreases in this case, while YOLOX-M and YOLOv8-M still maintain the highest mAP$_{50}$.


Fusion-based YOLOX-M and YOLOv8-M both get higher mAP$_{50\text{-}95}$ and mAP$_{50}$ than camera-based YOLOX-M and YOLOv8-M. Specifically, the fusion-based YOLOv8-M achieves an increase in mAP$_{50}$ from 84.4\% to 88.0\% compared to the camera-based YOLOv8-M. 
In adverse lighting and weather conditions, fusion-based models also achieve accuracy improvements. For example, in challenging lighting conditions, the fusion-based YOLOv8-M exhibits remarkable improvement, with the mAP$_{50}$ increasing from 74.8\% to 80.1\%, resulting in a noteworthy improvement of 5.3\% mAP$_{50}$.
Moreover, to enhance the density of radar point clouds, we perform experiments on accumulated 3-frame radar point clouds. It is explicit that denser radar point clouds are conducive to improving the mAP of object detection, both under normal conditions and adverse lighting and weather conditions. Despite our fusion network relying on basic operations derived from the camera model, the radar-camera fusion approaches still exhibit notable performance improvements. The highest observed improvement in performance amounts to 7.3\% mAP$_{50}$ for challenging lighting conditions.

\begin{figure}[htbp]
\centering
\subfloat[]{
\includegraphics[width=0.32\columnwidth]{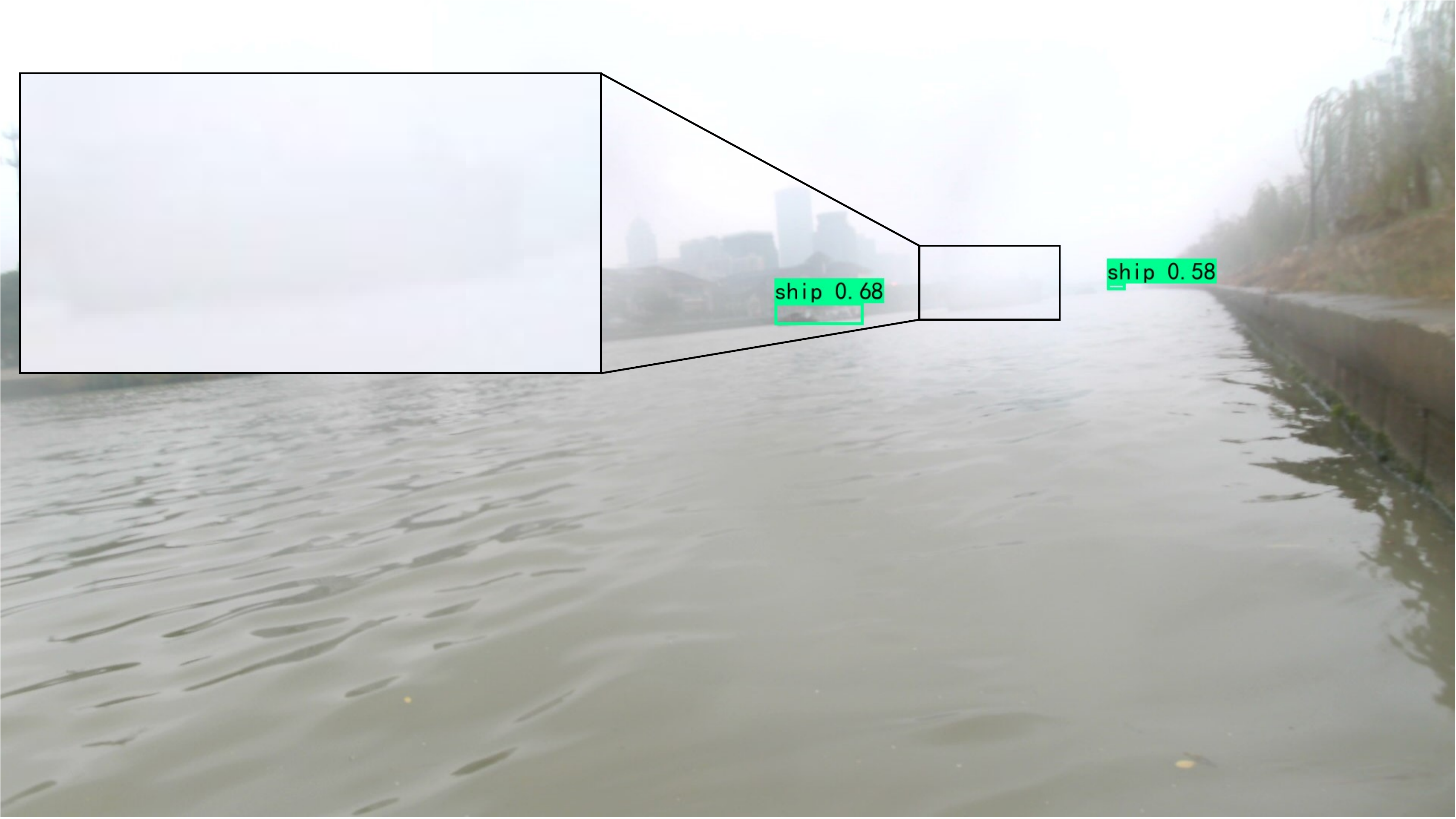}
\label{fig:det-1}
}
\quad
\hspace{-6.6mm}
\subfloat[]{
\includegraphics[width=0.32\columnwidth]{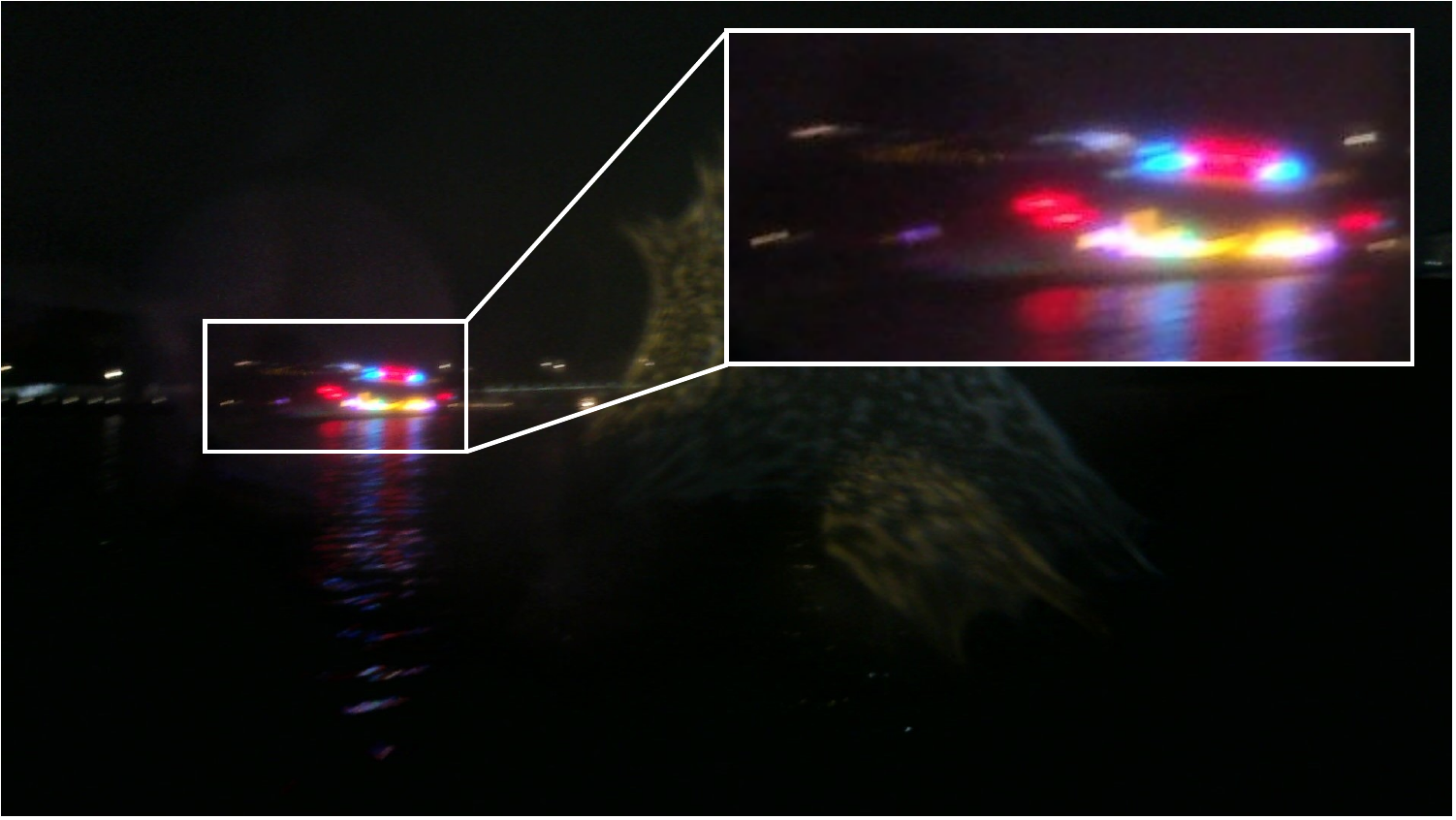}
\label{fig:det-2}
}
\quad
\hspace{-6.6mm}
\subfloat[]{
\includegraphics[width=0.32\columnwidth]{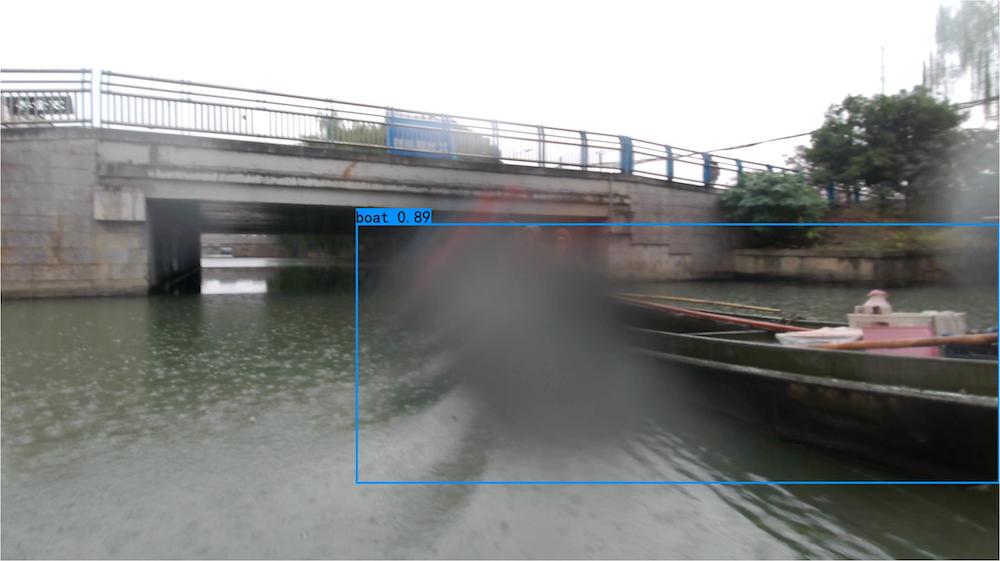}
\label{fig:det-3}
}
\vspace{-2mm}
\centering
\subfloat[]{
\includegraphics[width=0.32\columnwidth]{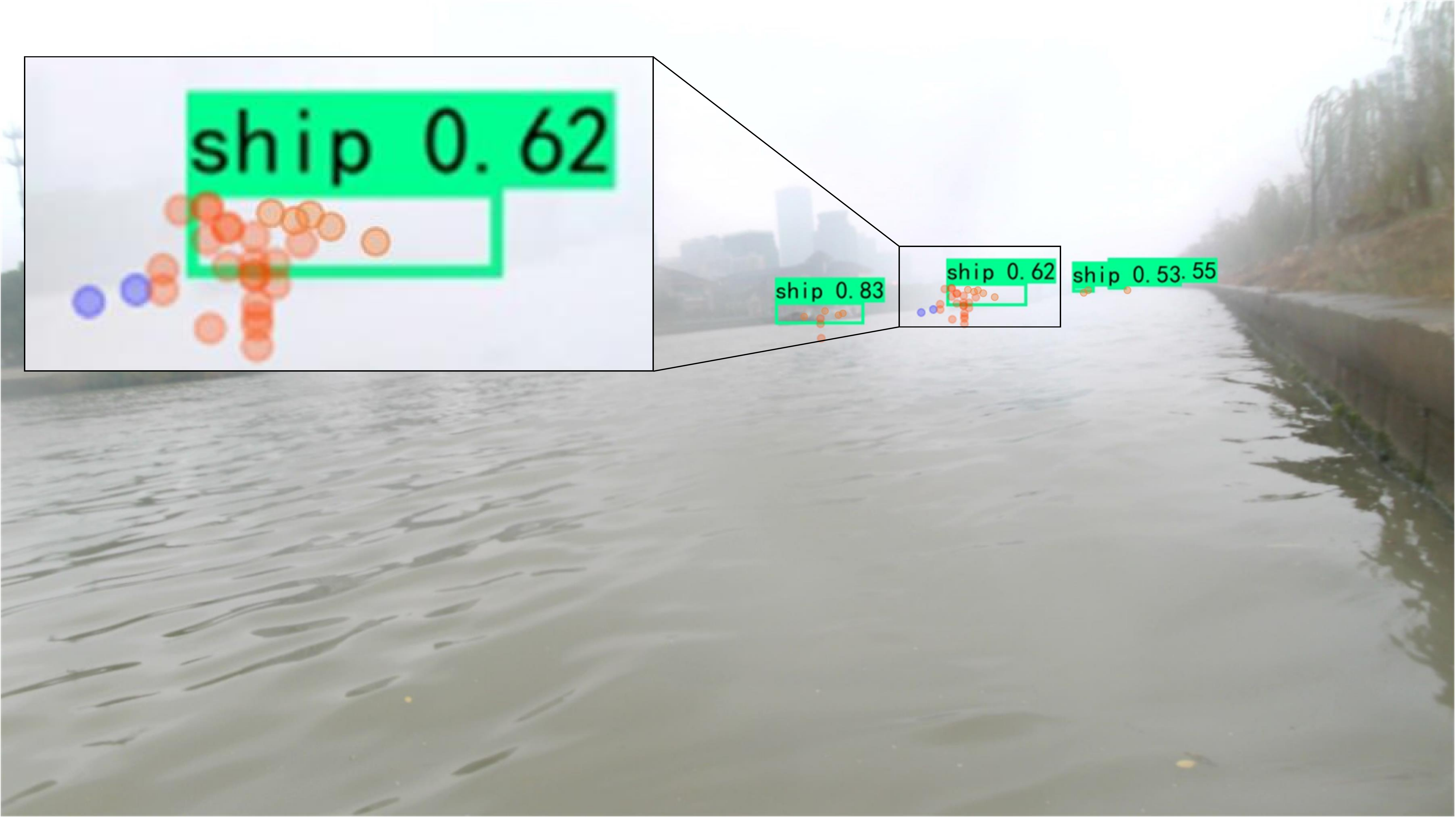}
\label{fig:det-4}
}
\quad
\hspace{-6.6mm}
\subfloat[]{
\includegraphics[width=0.32\columnwidth]{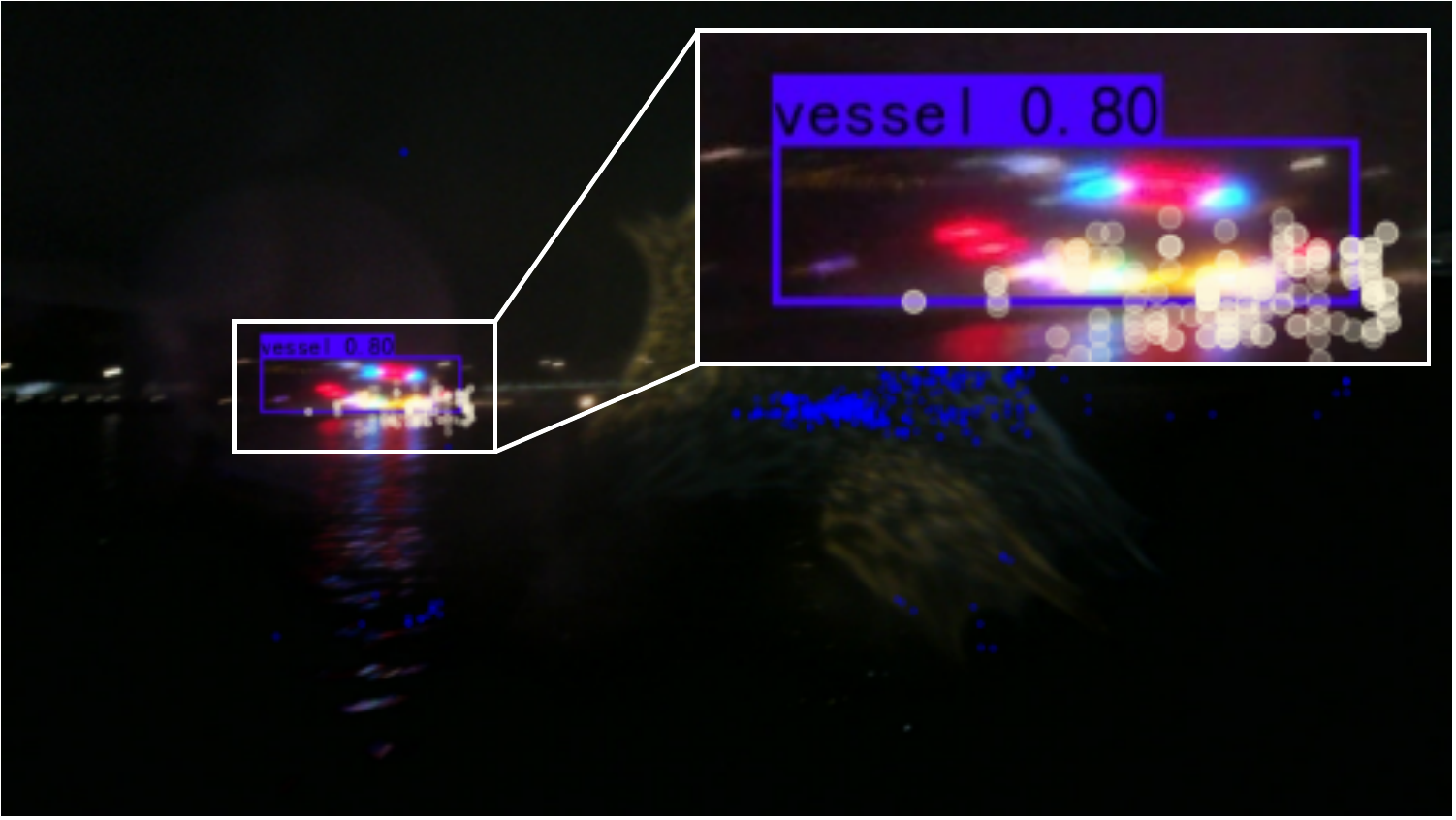}
\label{fig:det-5}
}
\quad
\hspace{-6.6mm}
\subfloat[]{
\includegraphics[width=0.32\columnwidth]{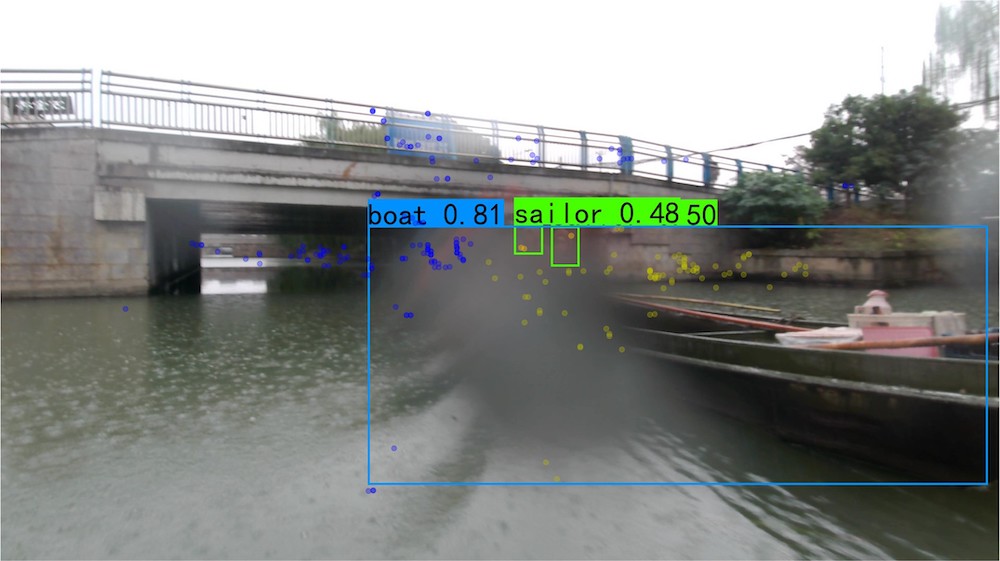}
\label{fig:det-6}
}
\caption{Visualization of object detection on WaterScenes under foggy weather (a, d), nighttime lighting (b, e) and partial sensor failure (c, f) conditions. The first row presents the detection results by the camera-based YOLOX-M. The second row presents the detection results by fusion-based YOLOX-M with input from camera and radar modalities.}
\label{fig:object-detection-figures}
\end{figure}

\textbf{Discussion.} Fig. \ref{fig:object-detection-figures} shows the representative outcomes obtained from both camera-based and fusion-based detection models.
Obviously, 4D radar enriches features to improve the recall of distant small objects (Fig. \subref*{fig:det-1} and Fig. \subref*{fig:det-4}), as well as objects located in dark environments (Fig. \subref*{fig:det-2} and Fig. \subref*{fig:det-5}). 
Additionally, due to the inherent unreliability of cameras, particularly with lens failure, as presented in Fig. \subref*{fig:det-3}, camera-based YOLOX-M fails to detect sailors on the boat. Fusion-based YOLOX-M successfully identifies the sailors, as shown in Fig. \subref*{fig:det-6}, thus improving the robustness of water surface perception. 
Although fusion-based models perform better than camera-based models, the confidence score is relatively low, and one sailor remains undetected. 


Designing efficient fusion methods based on the characteristics of different modalities is still a considerable challenge for water surfaces.
On the one hand, attention mechanisms for multi-modal fusion can be applied to the water surface domain. For example, the cross-attention modules in TransFusion \cite{bai2022transfusion} enable adaptive determination of what and where information should be taken from the camera and LiDAR data, leading to a robust and effective fusion strategy.
On the other hand, it is essential to address challenges specific to water surfaces. By leveraging techniques such as low-light enhancement \cite{zhou2022lednet}, waterdrop removal \cite{wen2023video}, rain and fog removal \cite{li2020all}, data quality from different modalities can be enhanced and contribute to more accurate fusion results.

\begin{table*}[!h]
\caption{Benchmark results of semantic segmentation on radar point clouds, including the Point Accuracy (PA) and mIoU of all classes. In the features column, $x, y, z$ denote the coordinates in the Cartesian system. $p, v, e$ denote reflected power, compensated Doppler velocity and elevation angle of the target, respectively.}
\center
\footnotesize
\begin{tabular}{ll|cc|cccccccc}
\toprule

\bf{Model} & \bf{Features} & \bf{PA} & \bf{mIoU} & \bf{Pier} & \bf{Buoy} & \bf{Sailor} & \bf{Ship} & \bf{Boat} & \bf{Vessel} & \bf{Kayak} & \bf{Clutter}  \\\midrule

PointMLP \cite{ma2022rethinking} & $x$, $y$, $z$ & 81.1 & 38.7 & 36.5 & 11.5 & 2.7 & 85.5 & 26.4 & 53.7 & 9.7 & 83.2\\
PointMLP \cite{ma2022rethinking} & $x$, $y$, $z$, $p$ & 86.3 & 51.7 & \textbf{61.2} & 18.3 & 2.9 & 87.9 & 50.7 & 55.2 & 6.8 & 86.8 \\
PointMLP \cite{ma2022rethinking} & $x$, $y$, $z$, $v$ & 83.0 &  45.3 & 45.4 & 37.6 & 3.2 & 90.6 & 41.2 & 51.7 & 4.8 & 87.5 \\
PointMLP \cite{ma2022rethinking} & $x$, $y$, $z$, $e$ & 84.1 &  46.9 & 44.8 & 33.5 & 0.7 & 86.9 & 41.4 & 59.0 & 24.2 & 84.5 \\
PointMLP \cite{ma2022rethinking} & $x$, $y$, $z$, $p$, $v$ & 86.9 & 53.0 & 50.4 & 48.3 & 1.1 & 92.1 & 54.3 & 81.8 & 7.8 & 88.0 \\
PointMLP \cite{ma2022rethinking} & $x$, $y$, $z$, $p$, $e$ & 87.1 & 53.5 & 56.8 & \textbf{51.9} & 1.1 & 90.5 & \textbf{59.5} & 80.1 & 0.6 & 87.3 \\
PointMLP \cite{ma2022rethinking} & $x$, $y$, $z$, $v$, $e$ & 84.7 & 49.7 & 48.7 & 32.3 & 1.3 & 87.0 & 41.1 & 60.1 & \textbf{43.2} & 84.2 \\
PointMLP \cite{ma2022rethinking} & $x$, $y$, $z$, $p$, $v$, $e$ & \textbf{89.7} & \textbf{55.7} & 45.7 & 39.8 & \textbf{8.3} & \textbf{93.2} & 57.8 & \textbf{88.6} & 21.1 & \textbf{90.7} \\
\midrule
Point-NN \cite{zhang2023parameter} & $x$, $y$, $z$, $p$, $v$, $e$ & 82.1 & 47.9 & 41.6 & 33.4 & 2.1 & 85.6 & 43.8 & 78.7 & 15.6 & 82.4 \\
PointNet++ \cite{qi2017pointnet++} & $x$, $y$, $z$, $p$, $v$, $e$ & 86.6 & 53.2 & 45.3 & 40.1 & 5.2 & 90.1 & 53.6 & 82.9 & 22.6 & 85.7 \\
Point Transformer \cite{zhao2021point} & $x$, $y$, $z$, $p$, $v$, $e$ & 87.9 & 54.4 & 42.0 & 37.8 & 8.0 & 92.1 & 58.1 & 87.6 & 20.7 & 88.9 \\
\bottomrule
\vspace{-2mm}
\end{tabular}

\label{tab:semantic_segmentation_PC}
\vspace{-2mm}
\end{table*}

\subsection{Radar Point Cloud Segmentation}

\textbf{Baseline.} We implement the semantic segmentation of radar point clouds based on different radar features. Table \ref{tab:semantic_segmentation_PC} indicates that PointMLP achieves the lowest PA and mIoU with only location features $x$, $y$ and $z$. 
By incorporating the physical features of the target, the combination of reflected power ($p$), compensated Doppler velocity ($v$), and elevation angle ($e$) achieves the highest accuracy, with 89.7\% PA and 55.7\% mIoU. 
Through ablation experiments, we discover that $p$, $v$, and $e$ all exhibit the potential to improve the semantic segmentation of radar point clouds.
Specifically, $p$ proves to be more effective in semantic segmentation than $v$ and $e$, as it serves as the reflected power indicating the materials of the target.


\begin{figure}[htbp]
\centering
\subfloat[]{
\includegraphics[width=0.32\columnwidth]{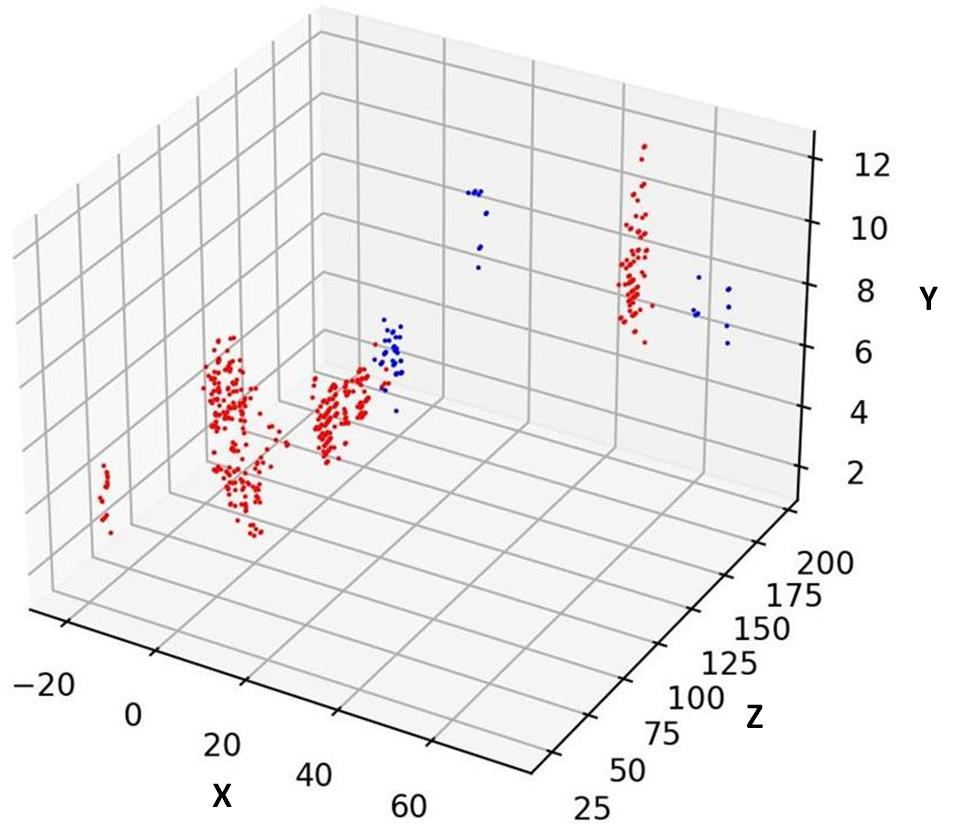}
\label{fig:radarPC-3Dseg1}
}
\quad
\hspace{-7mm}
\subfloat[]{
\includegraphics[width=0.32\columnwidth]{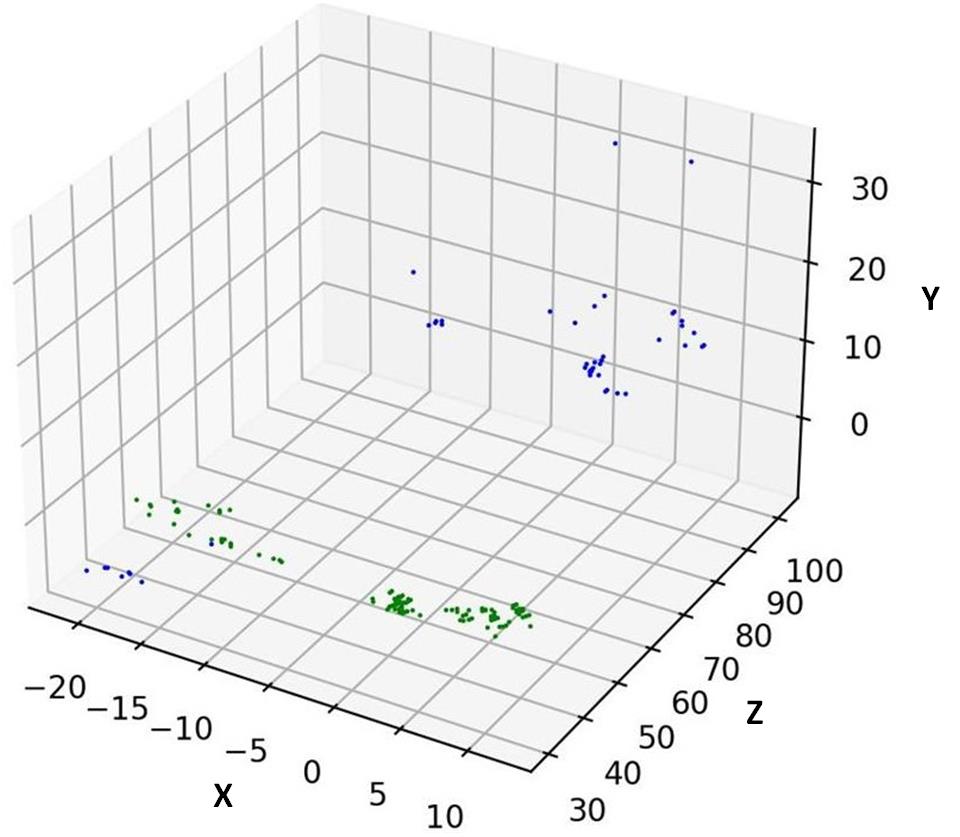}
\label{fig:radarPC-3Dseg2}
}
\quad
\hspace{-7mm}
\subfloat[]{
\includegraphics[width=0.32\columnwidth]{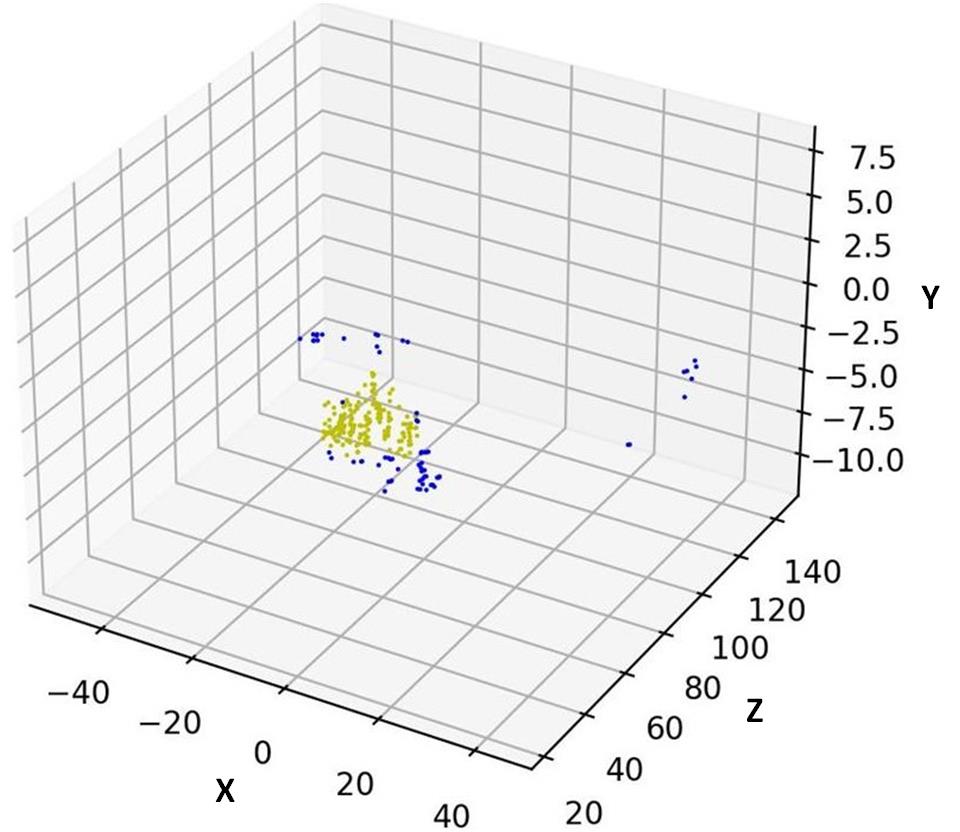}
\label{fig:radarPC-3Dseg3}
}
\vspace{-2mm}
\centering

\subfloat[]{
\includegraphics[width=0.32\columnwidth]{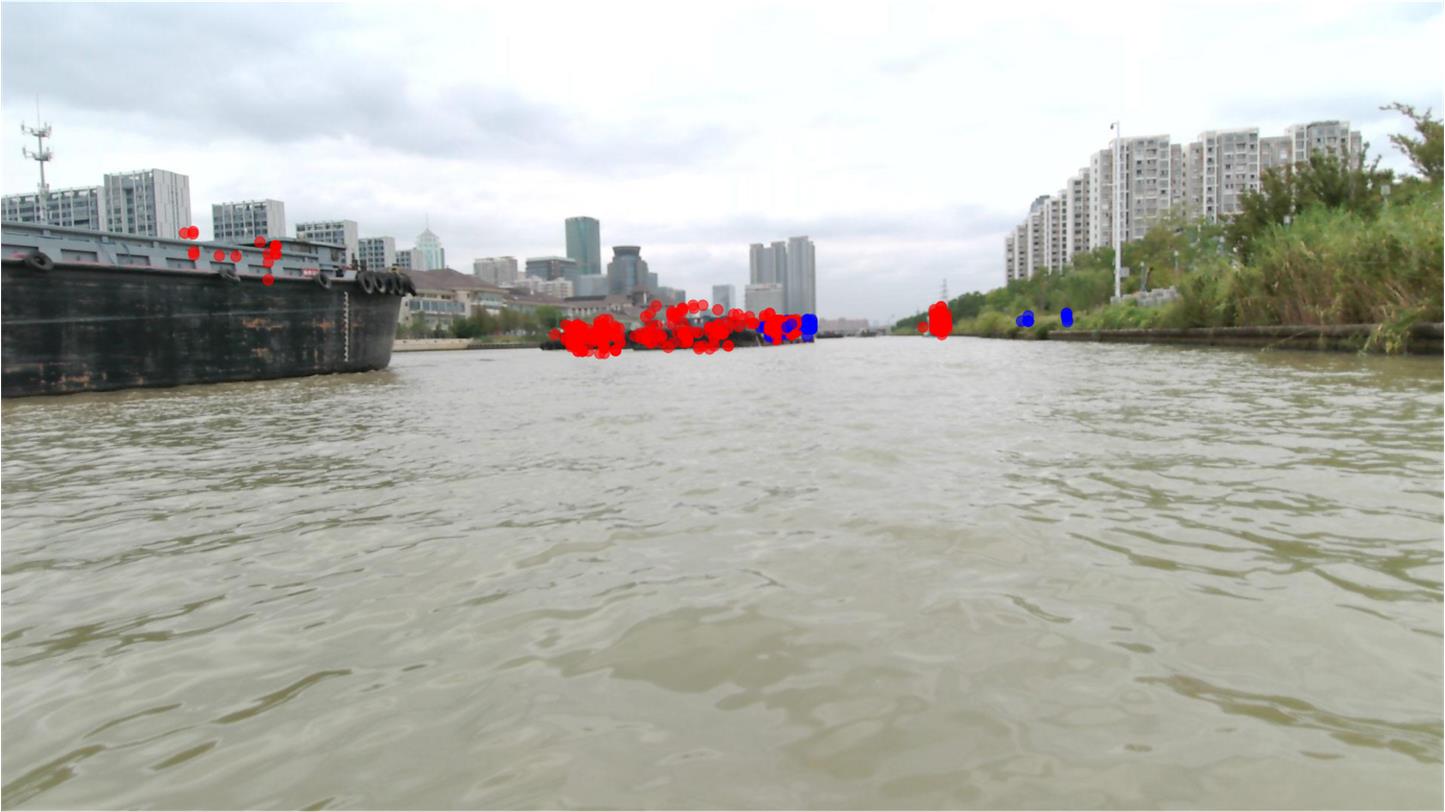}
\label{fig:radarPC-2Dseg1}
}
\quad
\hspace{-6.8mm}
\subfloat[]{
\includegraphics[width=0.32\columnwidth]{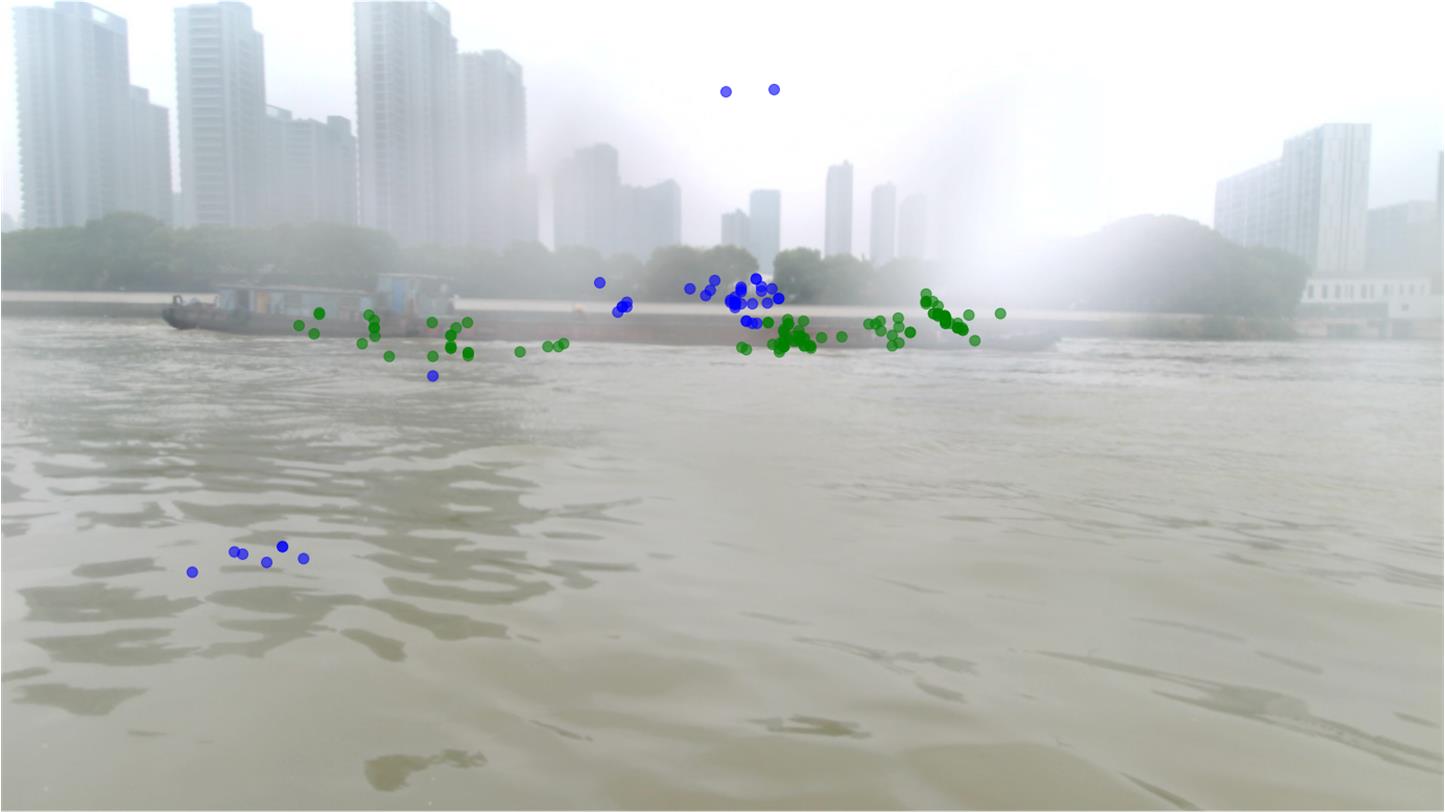}
\label{fig:radarPC-2Dseg2}
}
\quad
\hspace{-6.8mm}
\subfloat[]{
\includegraphics[width=0.32\columnwidth]{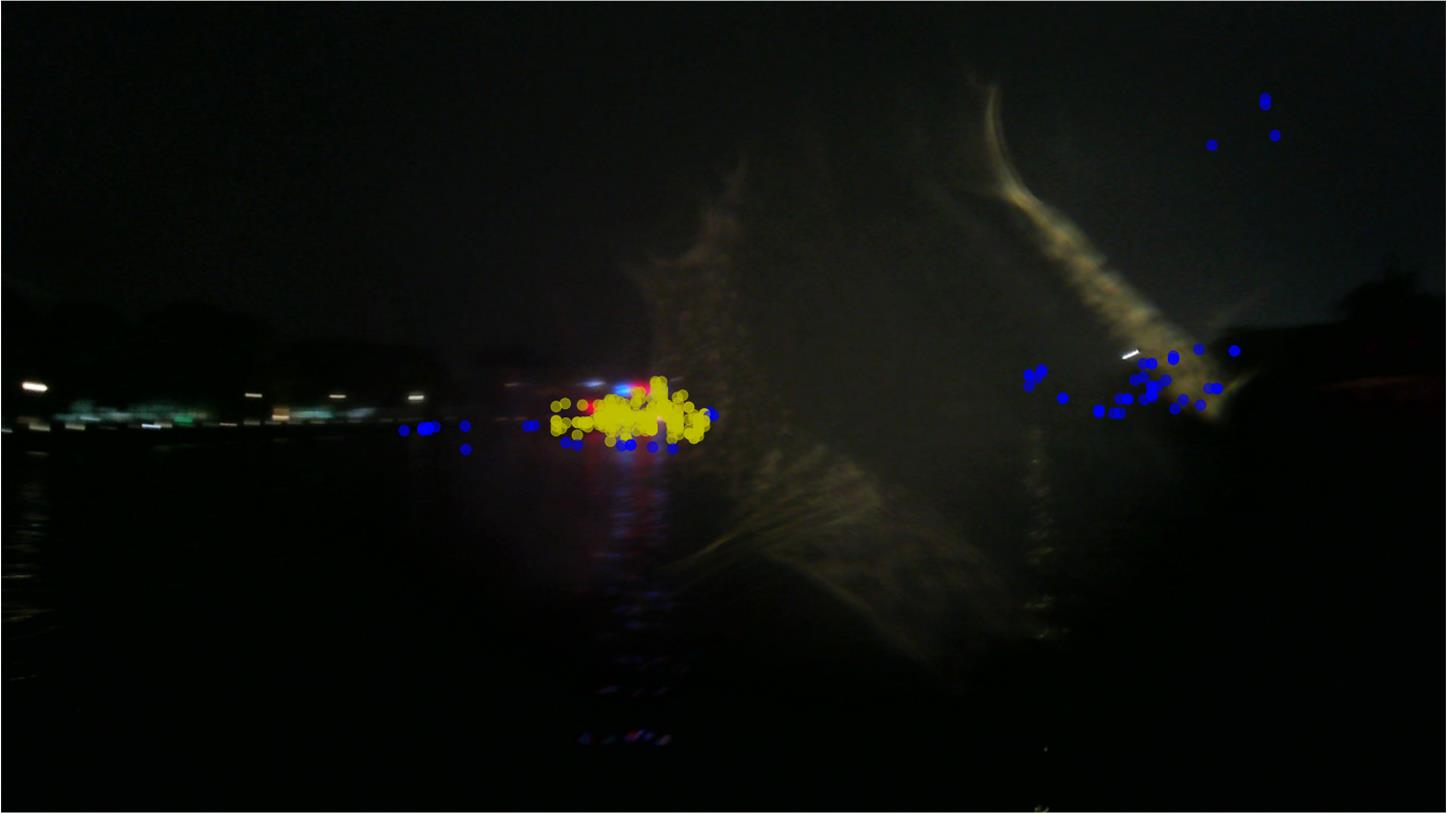}
\label{fig:radarPC-2Dseg3}
}
\caption{Visualization of radar point cloud semantic segmentation on WaterScenes. The first row is the semantic segmentation results of 3D radar point clouds in the world coordinates. The second row shows the radar point clouds projected onto the image plane. Blue point clouds indicate clutter while point clouds of other colors represent different kinds of objects.}
\label{fig:radar-pc-seg}
\end{figure}

\textbf{Discussion.} Fig. \ref{fig:radar-pc-seg} presents the visualization of 4D radar point cloud semantic segmentation in diverse environments, including normal weather, foggy weather and dark night. Radar point clouds demonstrate the capability to distinguish between targets and exhibit excellent robustness.
However, it is essential to note that unlike dense point clouds produced by LiDARs, radar point clouds are sparse and lack inherent semantic characteristics. Therefore, semantic segmentation of radar point clouds relies heavily on the physical attributes of the detected targets. 
Moreover, applying radar sensors on water surfaces may result in water clutter, thereby reducing the accuracy of point cloud segmentation. Thus, it is necessary to consider clutter removal methods such as those proposed in \cite{cheng2021new, cheng2021person} to enhance the segmentation accuracy of 4D radar point clouds on water surfaces.

\subsection{Camera Image Segmentation}

\begin{table}[htbp]
\caption{Benchmark results of semantic segmentation on WaterScenes.}
\center
\footnotesize
\begin{tabular*}{1\linewidth}{
p{2.2cm}<{}
p{1.1cm}<{\centering}
p{1.1cm}<{\centering}
p{1.1cm}<{\centering}
p{1.1cm}<{\centering}
}
\toprule
\bf{Model} & \bf{mIoU} & \bf{MPA} & \bf{OA}  & \bf{FPS} \\\midrule
DeepLabv3+ \cite{chen2018encoder} & 82.6 & 89.9 & 95.2 & \textbf{63.7} \\\midrule
HRNet \cite{wang2020deep} & 83.1 & 91.7 & 95.3 & 21.5\\\midrule
SegNeXt \cite{guo2022segnext} & 85.3 & 92.8 & 95.4 & 24.2 \\ \midrule
SegFormer \cite{xie2021segformer} & 85.7 & 93.1 & 95.4 & 59.5 \\ \midrule
Mask2Former \cite{cheng2022masked} & \textbf{86.6} & \textbf{93.9} & \textbf{96.2} & 6.8 \\
\bottomrule
\end{tabular*}
\vspace{1mm}
\label{tab:semantic_segmentation}
\end{table}

\begin{table}[htbp]
\caption{Benchmark results of instance segmentation on WaterScenes.}
\setlength\tabcolsep{5pt} 
\center
\footnotesize
\begin{tabular*}{1\linewidth}{p{2.2cm}<{}p{0.8cm}<{\centering}p{1.2cm}<{\centering}p{0.8cm}<{\centering}p{1.2cm}<{\centering}p{0.6cm}<{\centering}}
\toprule
\multicolumn{1}{l}{\multirow{2}[2]{*}{\textbf{Model}}} &
\multicolumn{2}{c}{\bf{Box}} & 
\multicolumn{2}{c}{\bf{Mask}} & \multicolumn{1}{c}{\multirow{2}[2]{*}{\textbf{FPS}}}
  \\ \cmidrule(lr){2-3}\cmidrule(lr){4-5}
\multicolumn{1}{c}{} & \bf{mAP$_{50}$} & \bf{mAP$_{50\text{-}95}$} & \bf{mAP$_{50}$} & \bf{mAP$_{50\text{-}95}$} 
\\\midrule
YOLACT \cite{bolya2019yolact} & 75.7 & 51.2 & 74.9 & 37.3 & 46.3 \\\midrule
SOLO \cite{wang2020solo} & - & - & 79.5 & 41.3 & 16.2 \\\midrule
YOLOv5-M \cite{yolov8} & 80.2 & 55.1 & 79.3 & 40.1 & 48.1\\\midrule
YOLOv8-M \cite{yolov8} & \textbf{85.9} & \textbf{58.2} & 79.2 & 44.8 & \textbf{54.6}\\\midrule
Mask2Former \cite{cheng2022masked} & - & - & \textbf{80.7} & \textbf{45.8} & 5.9 \\
\bottomrule
\end{tabular*}
\label{tab:instance-segmentation}
\end{table}

\textbf{Baseline of Semantic Segmentation.} Table \ref{tab:semantic_segmentation} presents that DeepLabv3+ obtains the highest FPS among the four models. Meanwhile, HRNet, using HRNetV1-W32 as its backbone, gets 83.1\% mIoU, 91.7\% MPA and 95.3\% OA. The above two models are based on pure-convolution networks. 
SegNeXt integrates the convolutional attention and employs MSCAN-B as the backbone, resulting in 95.4\% OA. 
SegFormer adopts multi-head self-attention at the last stage of its backbone and uses a naive MLP decoder, achieving the second highest 85.7\% mIoU among all models. As a transformer-based all-in-one segmentation model, Mask2Former achieves SOTA performance, exceeding SegFormer by 0.9\% mIoU.

\begin{table*}[!h]
\setlength\tabcolsep{5pt}
\caption{Benchmark results of panoptic perception on WaterScenes. In the Modalities column, C denotes the image modality from the camera sensor, and R denotes a single frame point cloud modality from the 4D radar sensor.}
\center
\footnotesize
\begin{tabular}{lccccccccccccccc}
\toprule
\multicolumn{1}{l}{\multirow{2}[2]{*}{\textbf{Model}}} & 
\multicolumn{1}{c}{\multirow{2}[2]{*}{\textbf{Modalities}}} &
\multicolumn{1}{c}{\multirow{2}[2]{*}{\textbf{Params (M)}}} &
\multicolumn{2}{c}{\bf{Object Detection}} &
\multicolumn{2}{c}{\bf{Free-Space Segmentation}} & 
\multicolumn{2}{c}{\bf{Waterline Segmentation}} & \multicolumn{1}{c}{\multirow{2}[2]{*}{\textbf{FPS}}}
  \\ \cmidrule(lr){4-5}\cmidrule(lr){6-7}\cmidrule(lr){8-9}
\multicolumn{3}{c}{} & \bf{mAP$_{50}$} & \bf{mAP$_{50\text{-}95}$} & \bf{OA} & \bf{mIoU} & \bf{OA} & \bf{mIoU}      
\\\midrule
YOLOP \cite{wu2022yolop} & C & 7.9 & 68.0 & 42.6 & 99.5 & 99.0 & 67.6 & 72.1 & 50.5 \\\midrule
HybridNets \cite{vu2022hybridnets} & C & 12.8 & 69.8 & 49.5 & 97.2 & 98.0 & 65.3 & 69.8 & 45.8\\\midrule
Achelous-MV-GDF-S0 \cite{guan2023achelous} & C + R & 1.6 & 81.1 & 51.0 & 99.6 & 99.3 & 68.3 & 65.0 & \bf{70.3} \\\midrule
Achelous-MV-GDF-S1 \cite{guan2023achelous} & C + R & 2.8 & 83.5 & 54.1 & 99.6 & 99.4 & 69.5 & 68.7 & 69.6  \\\midrule
Achelous-MV-GDF-S2 \cite{guan2023achelous} & C + R & 5.3 & \bf{85.5} & \bf{56.0} & \bf{99.7} & \bf{99.6} & \bf{70.3} & \bf{72.2} & 68.5  \\\midrule
\end{tabular}
\label{tab:multi-task}
\end{table*}

\textbf{Baseline of Instance Segmentation.} 
Experiments show that YOLOv8-M outperforms all other box-based models with 85.9\% mAP$_{\text{50}}$, 58.2\% mAP$_{50\text{-}95}$, and 54.6 FPS in Table \ref{tab:instance-segmentation}. 
Transformer-based Mask2Former achieves the highest mask mAP$_{\text{50}}$ of 80.7\% and mAP$_{50\text{-}95}$ of 45.8\%.
For CNN-based networks, SOLO obtains the highest mask mAP$_{\text{50}}$ of 79.5\%. 
Overall, YOLOv8-M offers an excellent trade-off between accuracy and inference speed.

\begin{figure}[h]
\centering
\subfloat[]{
\includegraphics[width=0.32\columnwidth]{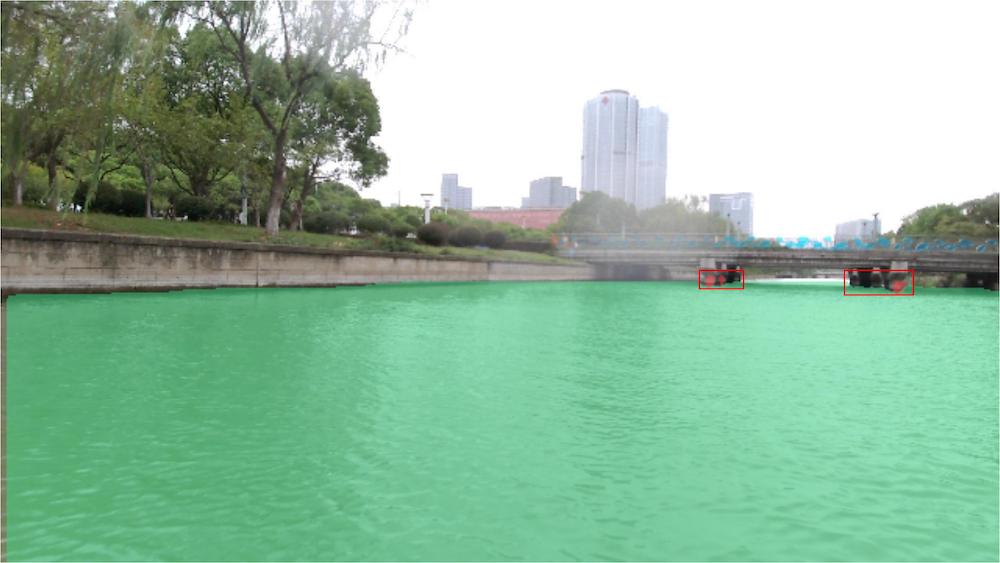}
\label{fig:semantic-1}
}
\quad
\hspace{-6.6mm}
\subfloat[]{
\includegraphics[width=0.32\columnwidth]{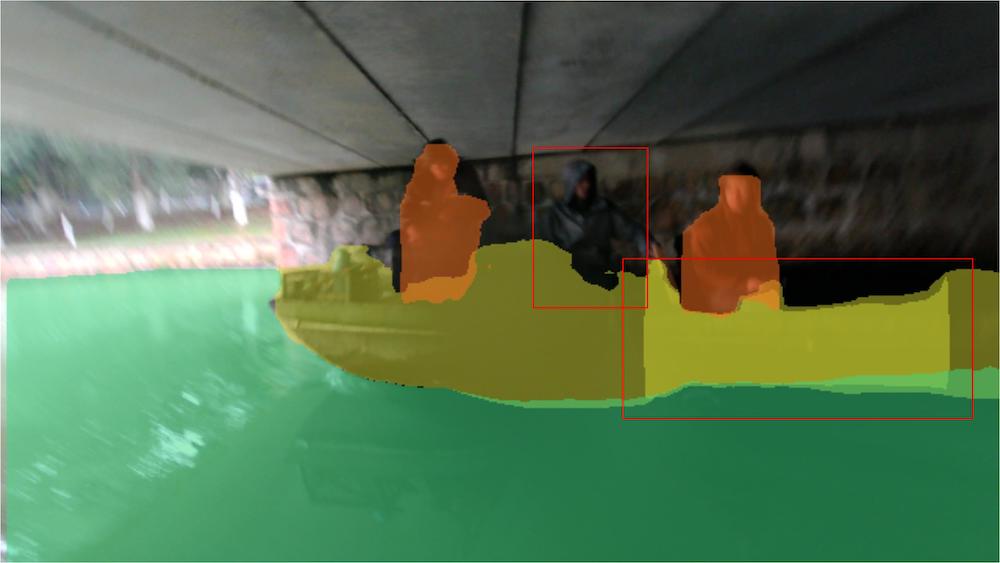}
\label{fig:semantic-2}
}
\quad
\hspace{-6.6mm}
\subfloat[]{
\includegraphics[width=0.32\columnwidth]{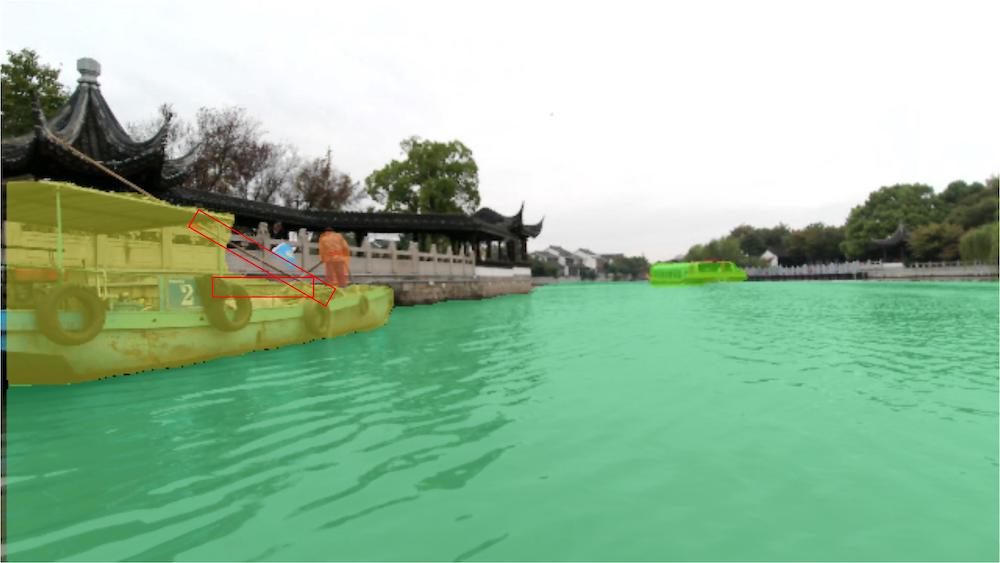}
\label{fig:semantic-3}
}
\vspace{-2mm}

\centering
\subfloat[]{
\includegraphics[width=0.32\columnwidth]{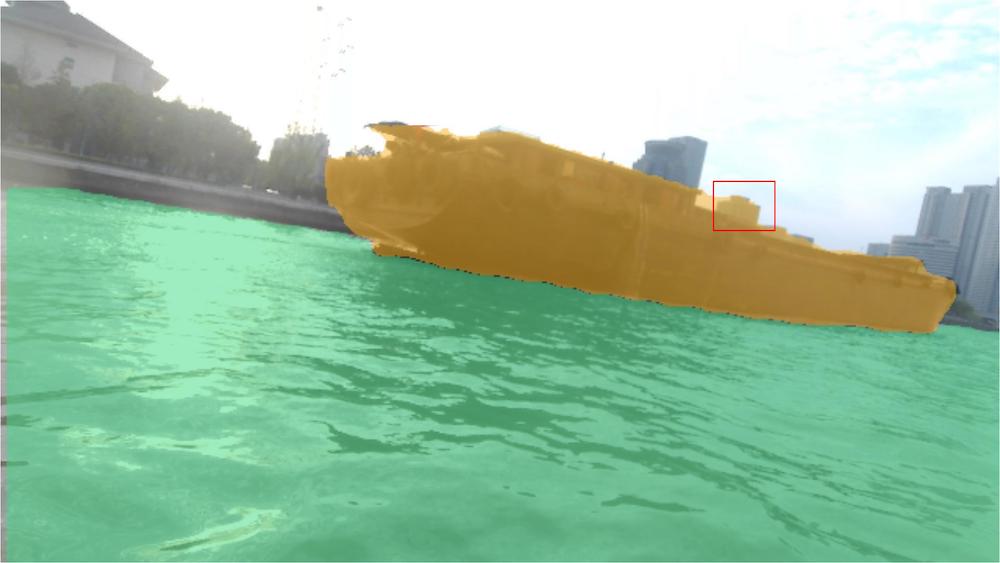}
\label{fig:semantic-4}
}
\quad
\hspace{-6.6mm}
\subfloat[]{
\includegraphics[width=0.32\columnwidth]{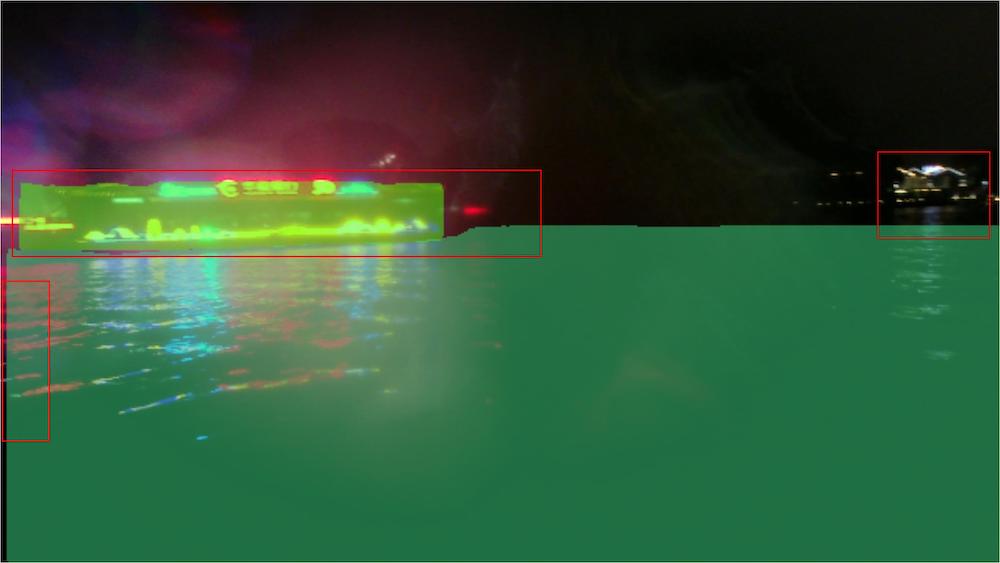}
\label{fig:semantic-5}
}
\quad
\hspace{-6.6mm}
\subfloat[]{
\includegraphics[width=0.32\columnwidth]{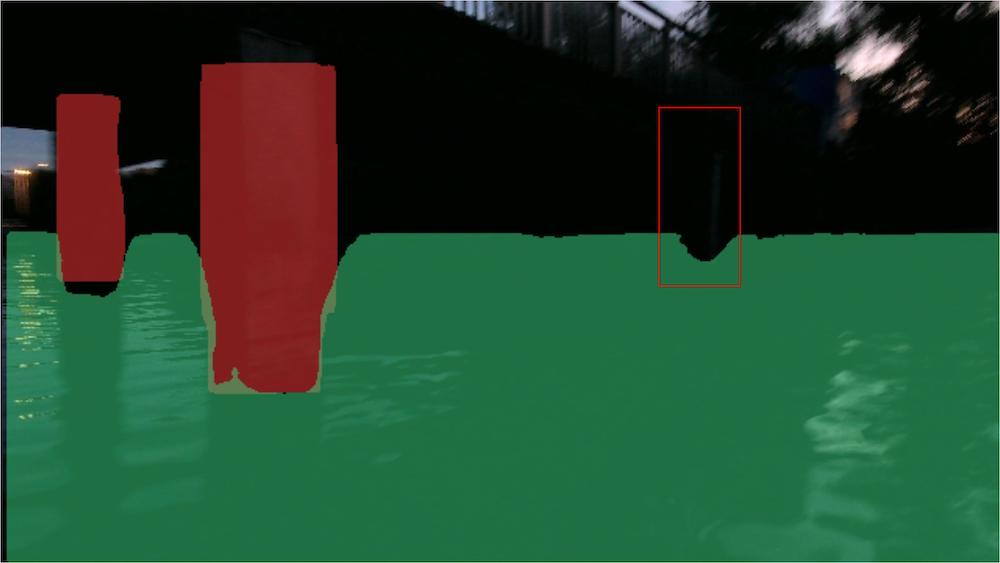}
\label{fig:semantic-6}
}
\caption{Visualization of semantic segmentation on WaterScenes. (a) Blurred segmentation of distant piers. (b) Fuzzy sailor-boat boundaries. (c) Ambiguous complex boat segmentation. (d) Buildings misclassified as ship parts. (e) Indistinct ship edges in low light. (f) Piers excluded from segmentation.}
\label{fig:semantic}
\end{figure}

\textbf{Discussion.} As illustrated in Fig. \ref{fig:semantic}, our WaterScenes presents considerable challenges. 
First of all, Fig. \subref*{fig:semantic-1} and Fig. \subref*{fig:semantic-2} demonstrate that models are not good at segmenting small objects (e.g., piers) and objects that are in close contact, such as sailors and boats.
Additionally, Fig. \subref*{fig:semantic-3} suggests that models struggle with boats that have complex structures (e.g., a boat with a roof supported by poles), especially when sailors are present on the boat.
Furthermore, background buildings are sometimes misidentified as part of the same object as the ship with the steel structure, as shown in Fig. \subref*{fig:semantic-4}.
In the case of dim lighting conditions, the segmentation results become quite rough or even completely missing, as illustrated in Fig. \subref*{fig:semantic-5} and Fig. \subref*{fig:semantic-6}.
Inaccurate segmentation of objects poses a significant challenge to the autonomous driving of USVs. 
Consequently, in addition to specific network design for camera modality on water surfaces, leveraging radar to assist camera image segmentation is a valuable research direction.


\subsection{Panoptic Perception}

\textbf{Baseline.} As can be seen from Table \ref{tab:multi-task}, benchmark results indicate both the feasibility of our dataset for panoptic perception and the challenges associated with multi-task perception on water surfaces.
In general, fusion-based Achelous exhibits superior performance compared to camera-based YOLOP and HybridNets among object detection, free-space segmentation and waterline segmentation tasks. In terms of object detection, Achelous outperforms HybridNets by 15.7\% mAP$_{50}$, demonstrating the effectiveness of radar-camera fusion on water surfaces.  
However, it still has a lower detection mAP than the YOLOv8-M model with radar-camera fusion in Table \ref{tab:object-detection-baselines}, which is specifically designed for the object detection task. 

\begin{figure*}
\centering
\subfloat[]{
\includegraphics[width=0.39\columnwidth]{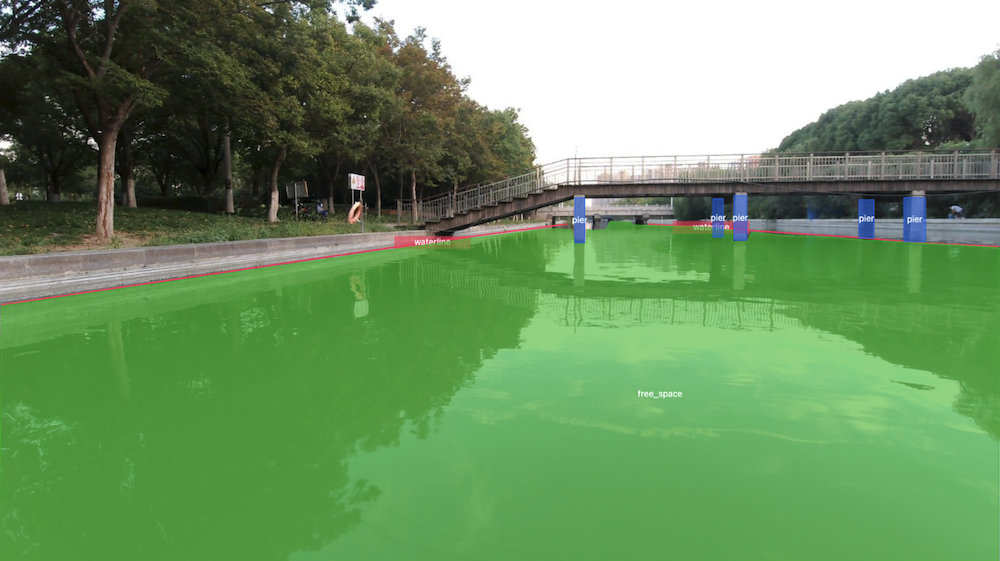}
\label{fig:GT-1}
}
\hspace{-6.6mm}
\quad
\subfloat[]{
\includegraphics[width=0.39\columnwidth]{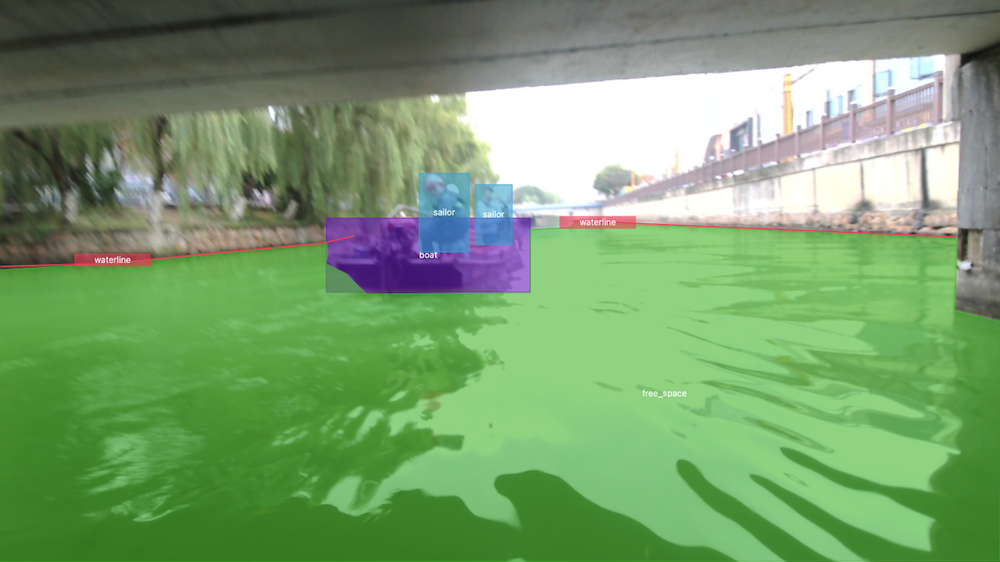}
\label{fig:GT-2}
}
\hspace{-6.6mm}
\quad
\subfloat[]{
\includegraphics[width=0.39\columnwidth]{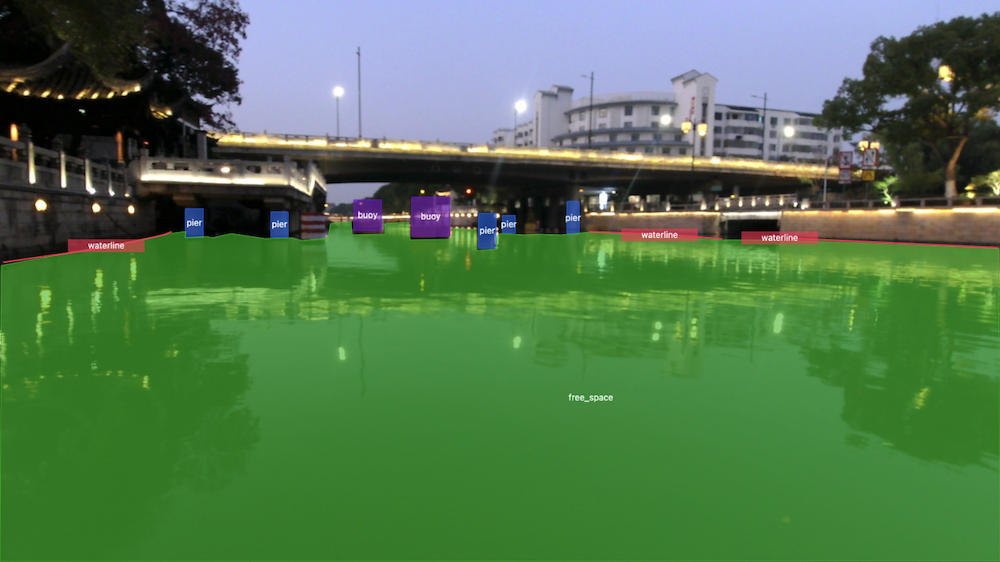}
\label{fig:GT-3}
}
\hspace{-6.6mm}
\quad
\centering
\subfloat[]{
\includegraphics[width=0.39\columnwidth]{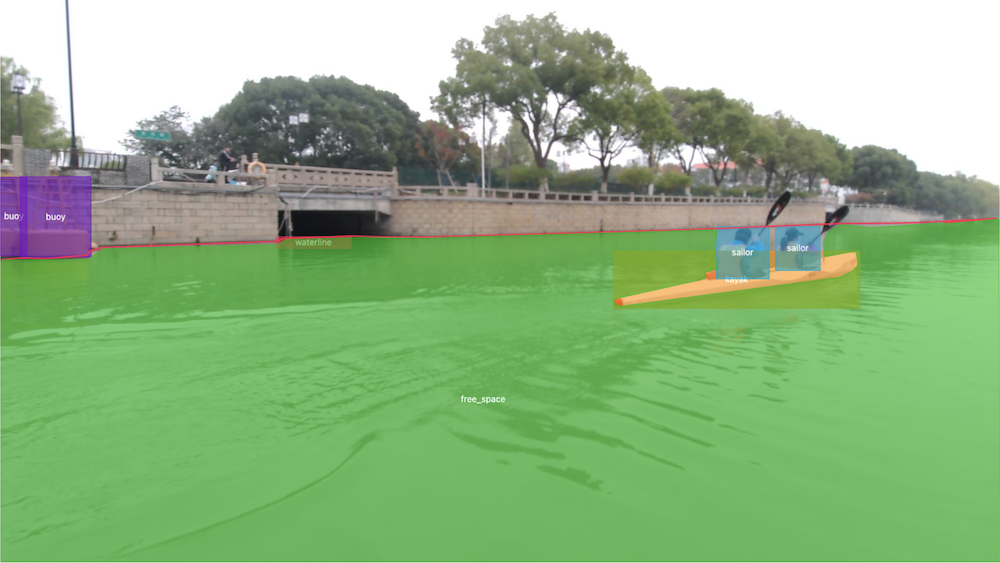}
\label{fig:GT-4}
}
\hspace{-6.6mm}
\quad
\subfloat[]{
\includegraphics[width=0.39\columnwidth]{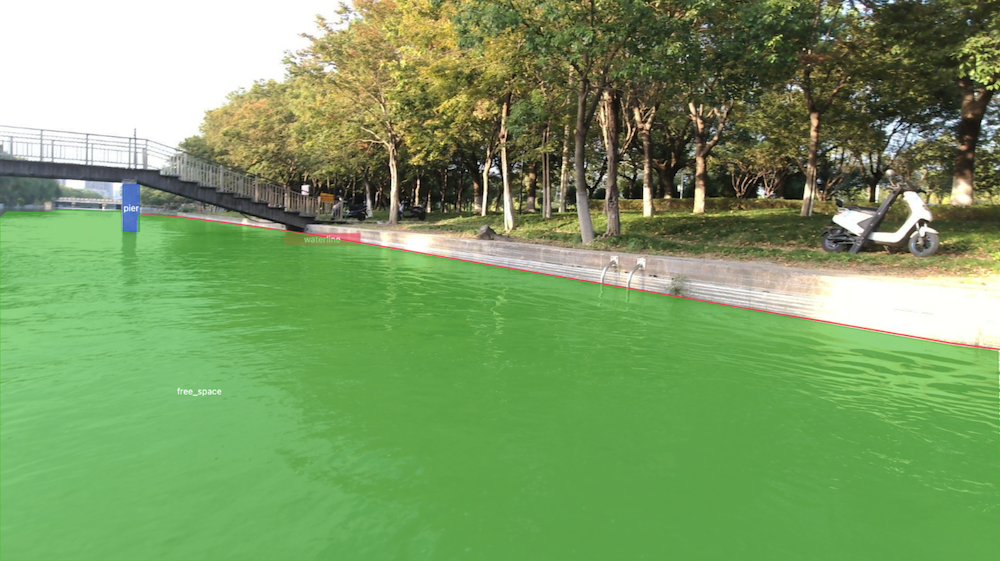}
\label{fig:GT-5}
}

\subfloat[]{
\includegraphics[width=0.39\columnwidth]{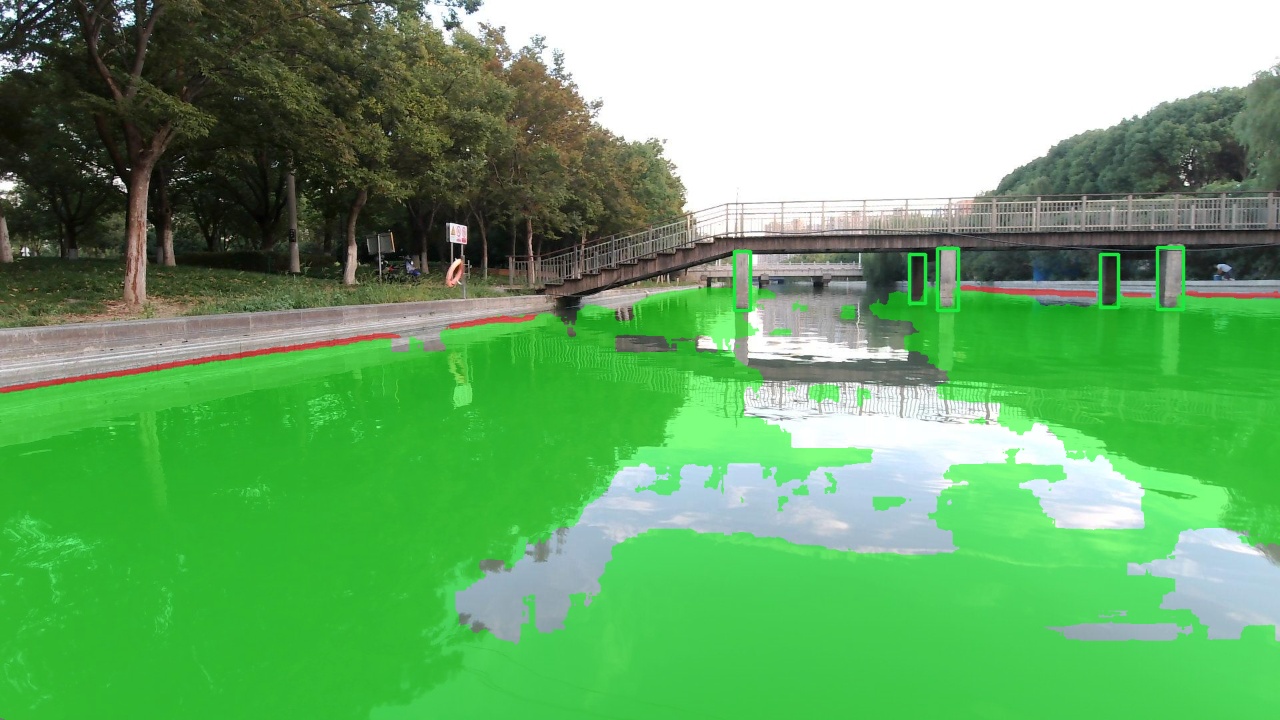}
\label{fig:yolop-1}
}
\hspace{-6.6mm}
\quad
\subfloat[]{
\includegraphics[width=0.39\columnwidth]{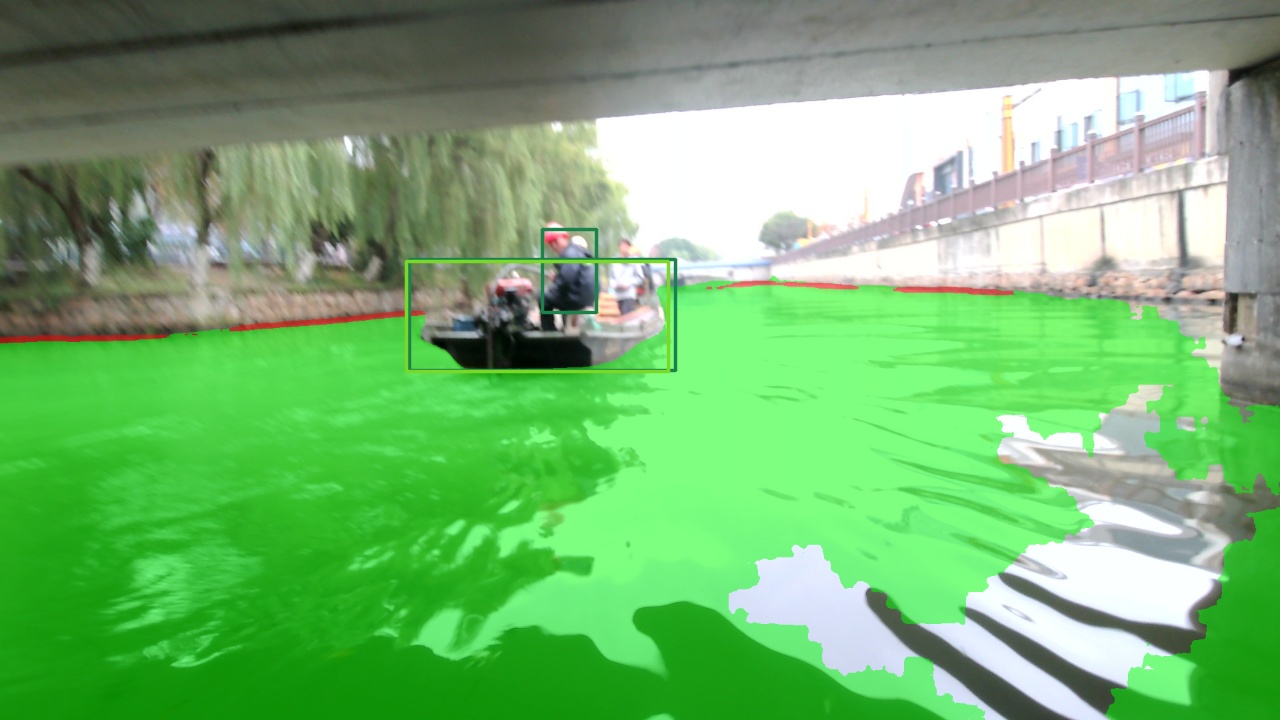}
\label{fig:yolop-2}
}
\hspace{-6.6mm}
\quad
\subfloat[]{
\includegraphics[width=0.39\columnwidth]{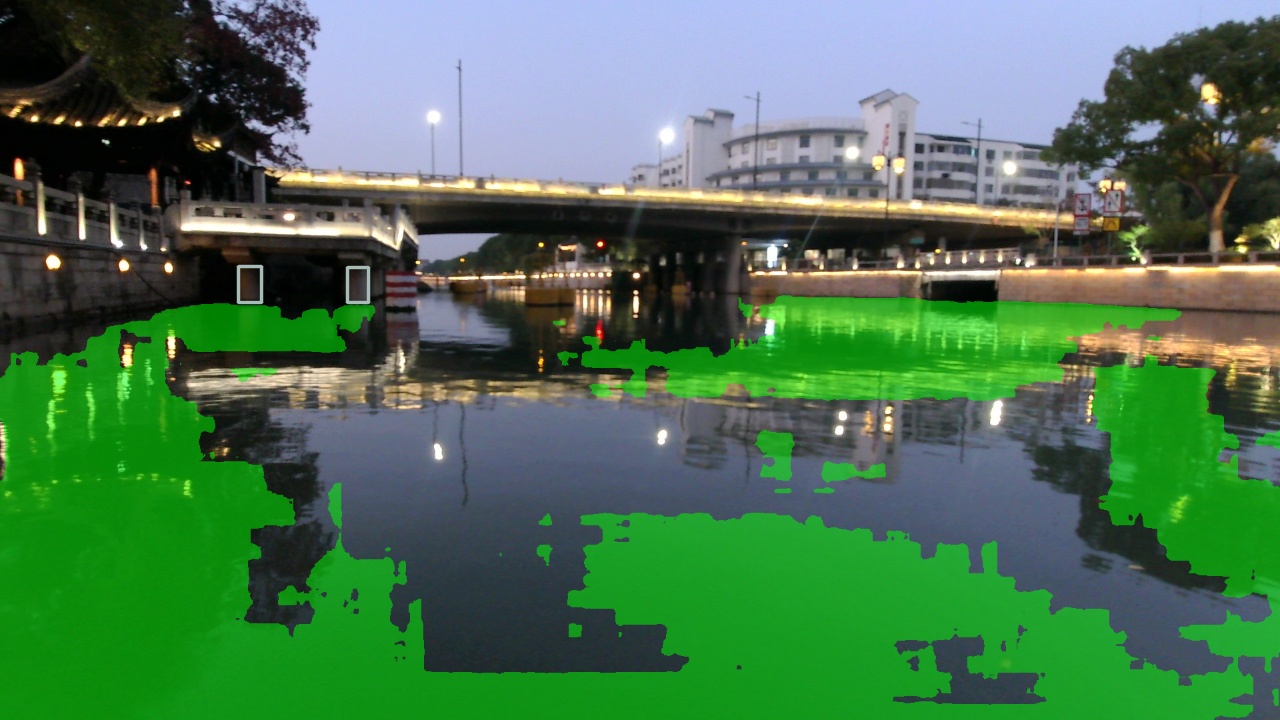}
\label{fig:yolop-3}
}
\hspace{-6.6mm}
\quad
\centering
\subfloat[]{
\includegraphics[width=0.39\columnwidth]{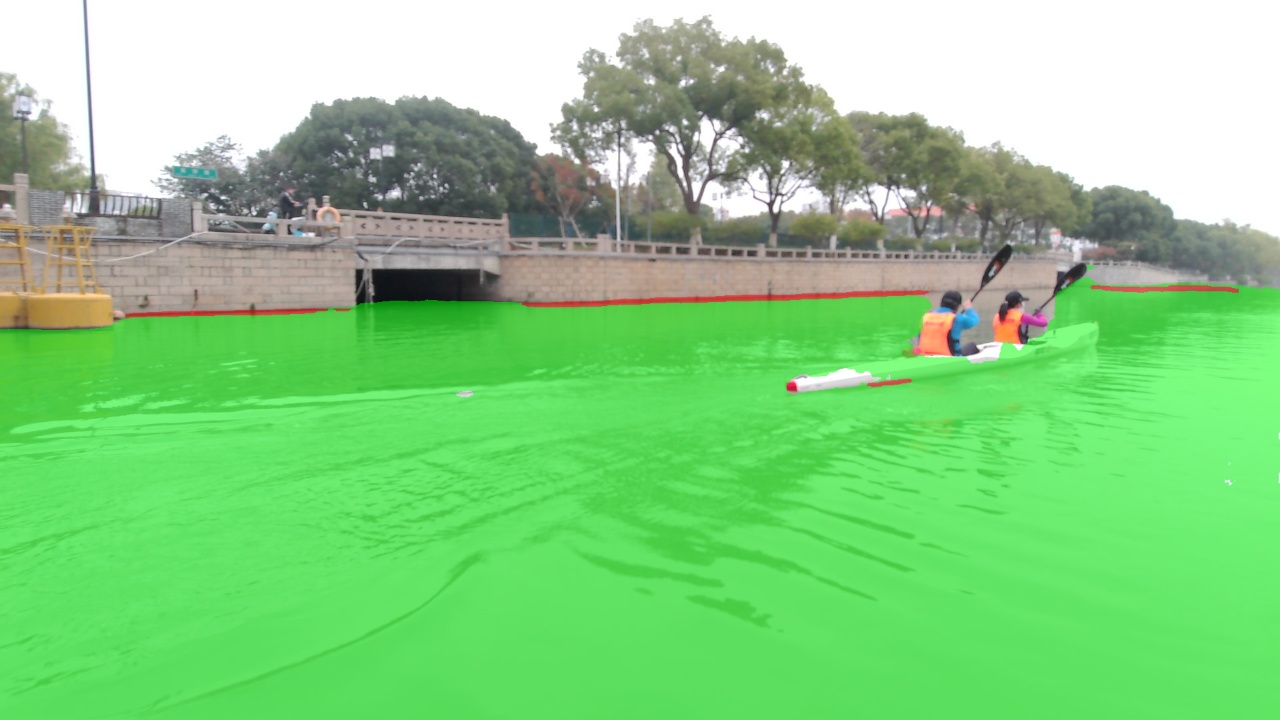}
\label{fig:yolop-4}
}
\hspace{-6.6mm}
\quad
\subfloat[]{
\includegraphics[width=0.39\columnwidth]{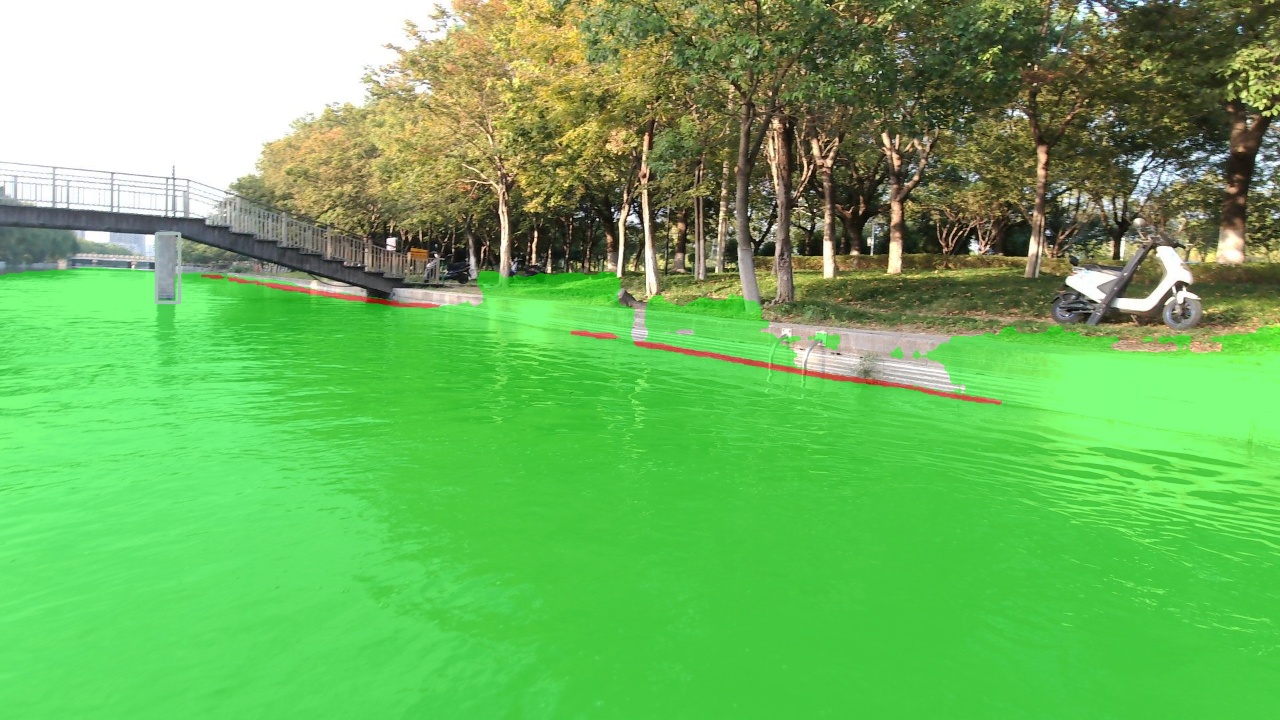}
\label{fig:yolop-5}
}

\centering
\subfloat[]{
\includegraphics[width=0.39\columnwidth]{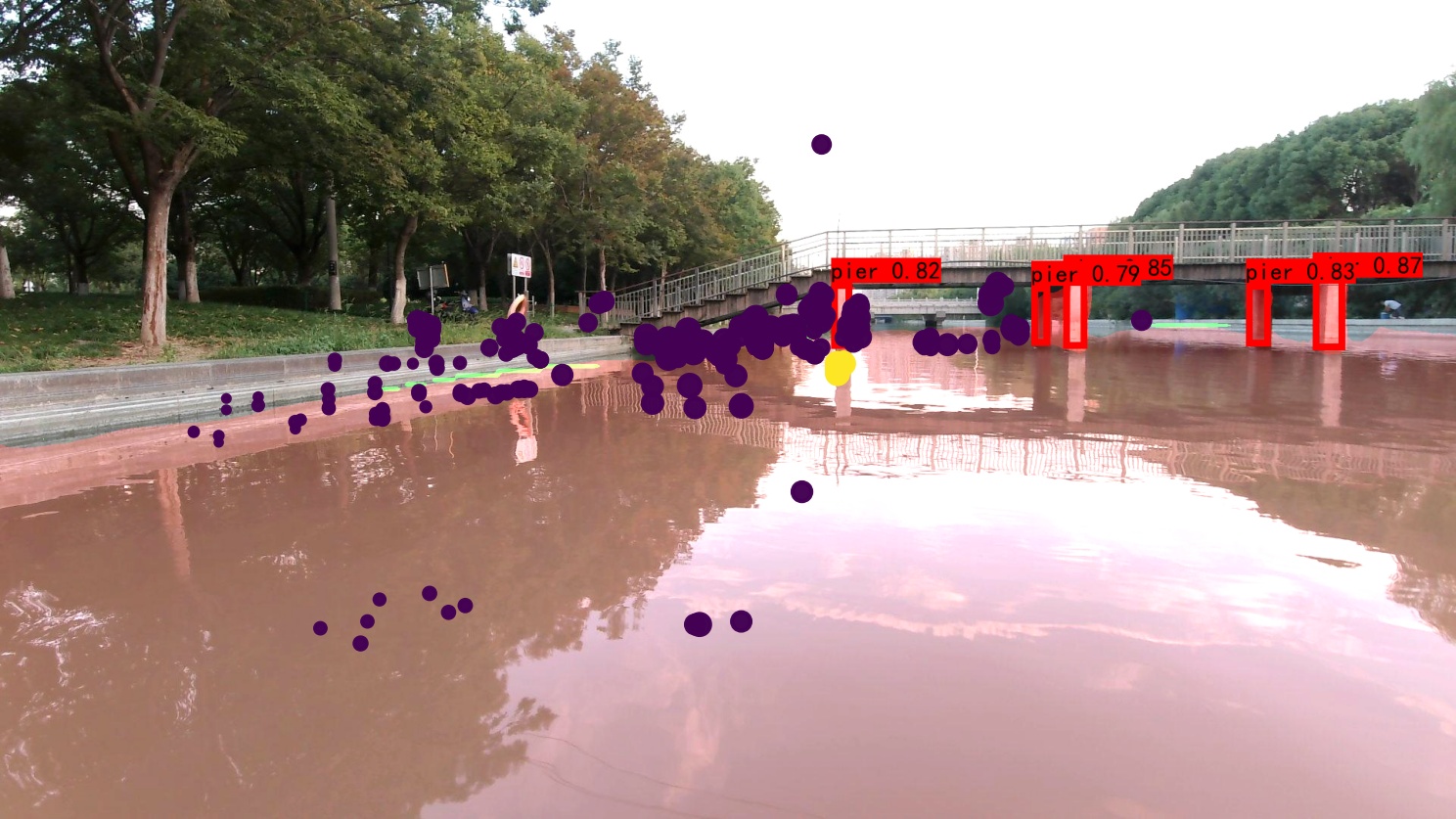}
\label{fig:achelous-1}
}
\hspace{-6.6mm}
\quad
\subfloat[]{
\includegraphics[width=0.39\columnwidth]{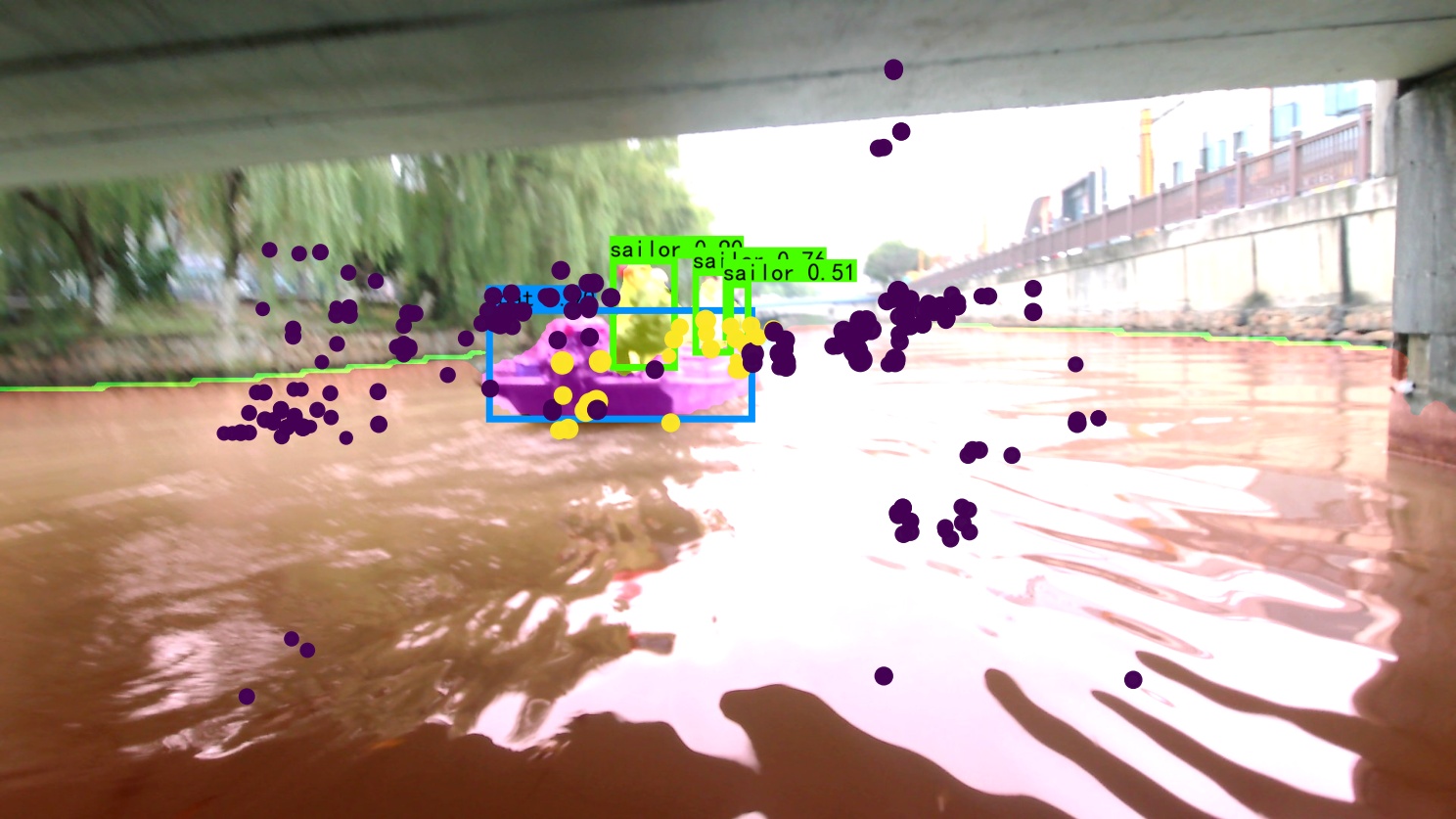}
\label{fig:achelous-2}
}
\hspace{-6.6mm}
\quad
\subfloat[]{
\includegraphics[width=0.39\columnwidth]{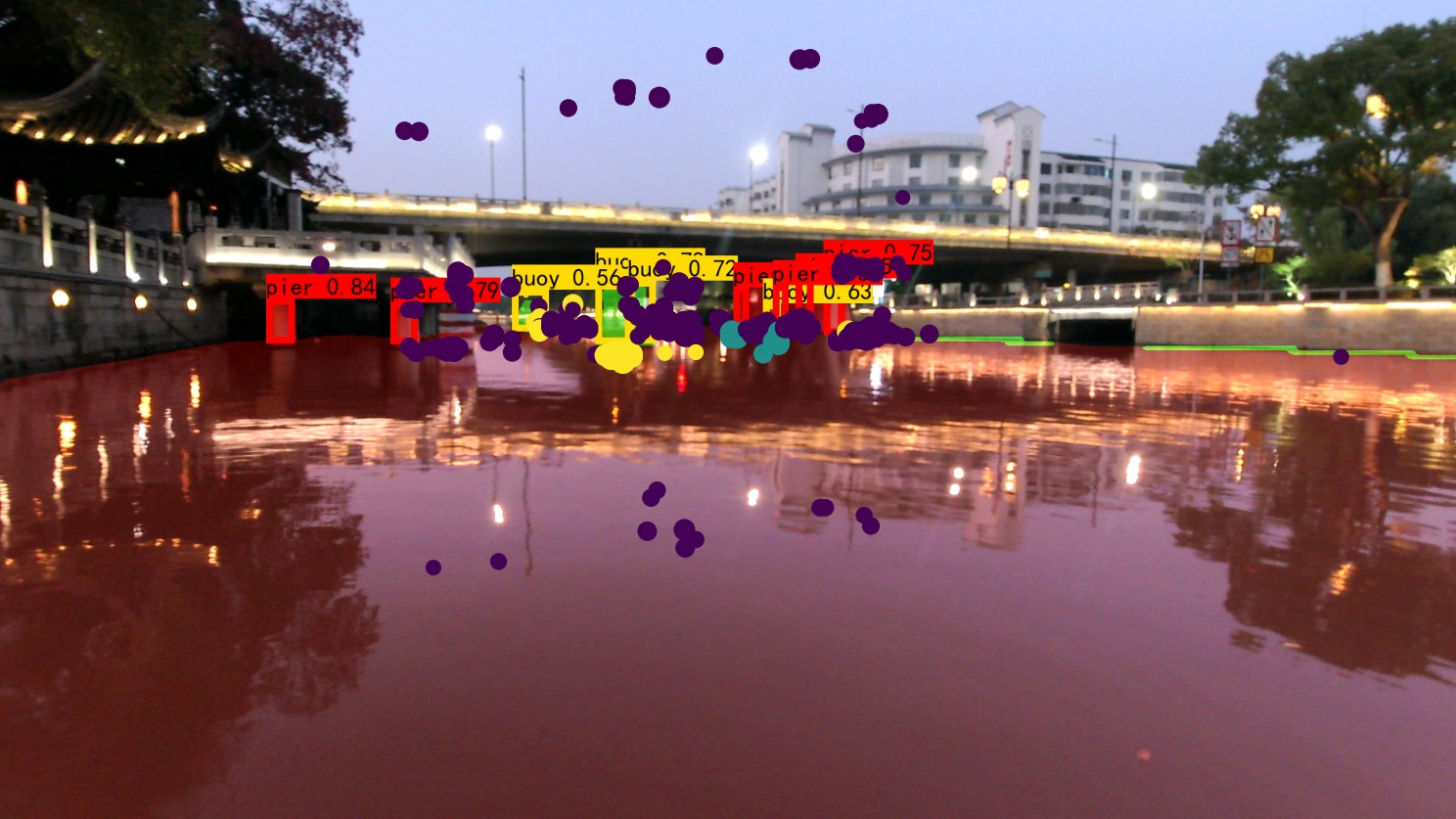}
\label{fig:achelous-3}
}
\hspace{-6.6mm}
\quad
\centering
\subfloat[]{
\includegraphics[width=0.39\columnwidth]{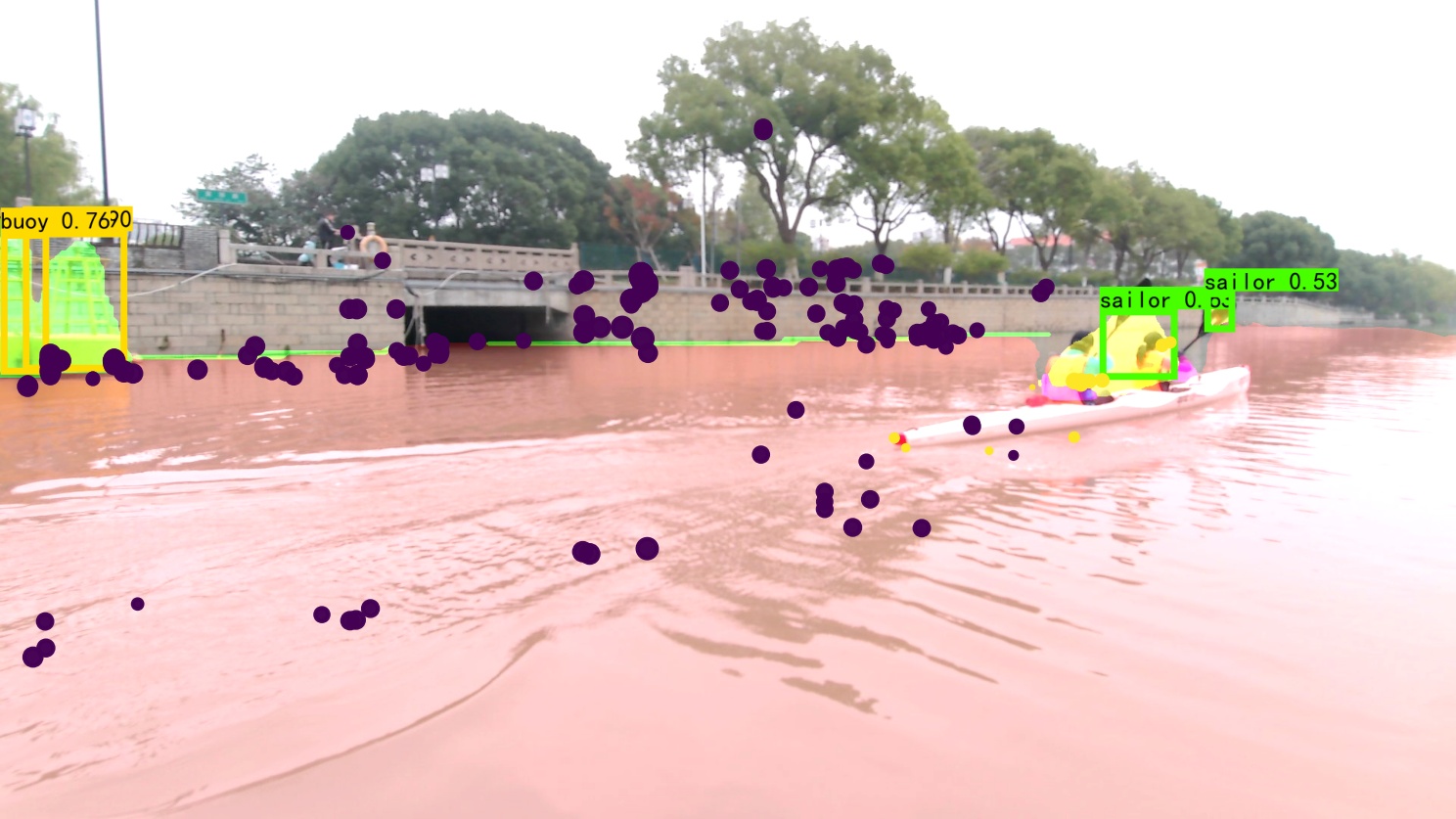}
\label{fig:achelous-4}
}
\hspace{-6.6mm}
\quad
\subfloat[]{
\includegraphics[width=0.39\columnwidth]{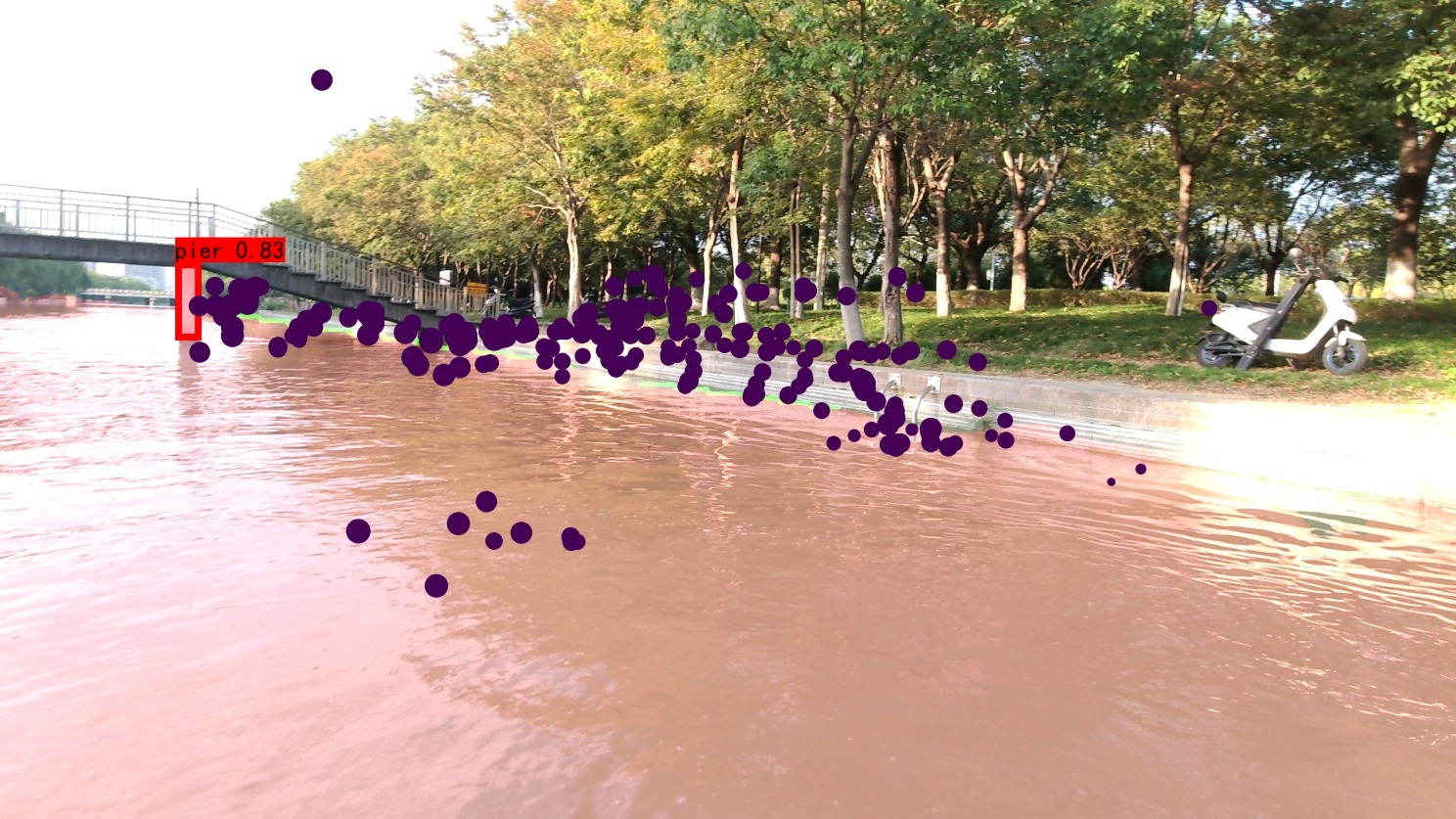}
\label{fig:achelous-5}
}

\caption{Visualization of panoptic perception on WaterScenes. Images in the first row are the ground truth, in the second row are results of camera-based YOLOP, and in the third row are results of fusion-based Achelous. Panoptic perception includes object detection (boxes), free-space segmentation (masks) and waterline segmentation (lines).}
\label{fig:multi-task}
\end{figure*}

\textbf{Discussion.} Unlike panoptic perception of object detection, drivable area and lane line segmentation for autonomous vehicles on roads, reflections on water surfaces and the unclear boundary line between water and shore make it challenging to segment free-space and waterlines. 
For example, as shown in Fig. \subref*{fig:GT-1} and \subref*{fig:GT-2}, areas of bright spots caused by light and waves are incorrectly identified as free-space. 
The water surface mirrors buildings on the shore at night, further complicating the segmentation of the free-space area, as demonstrated in Fig. \subref*{fig:GT-3}.
Small objects tend to have less contact area with the water surface, making them easier to be missed, as is indicated in Fig. \subref*{fig:GT-4}. 
Furthermore, as illustrated in Fig. \subref*{fig:GT-5}, the boundary between the water surface and the shore is unclear, especially in low-light environments. Consequently, the model misidentifies the waterline as part of the water surface, presenting a potential risk of collision in real-world scenarios.



%
%

Multi-modal panoptic perception on water surfaces remains an unexplored and valuable research direction. In Multi-Task Learning (MTL) paradigm, multiple task-specific heads share the feature extraction process. The co-training strategy across tasks could leverage feature abstraction to save computation cost for onboard chips. Panoptic perception also serves for downstream tasks on water surfaces, such as path planning, obstacle avoidance and navigation control for USVs. Therefore, lightweight architectures that can handle multiple modalities and multiple tasks in real-time are highly desirable for edge devices on USVs.

\section{Discussions}
\label{sec:Discussions}

\subsection{Dataset Diversity}


\begin{table}[htbp]
\caption{Experiments of dataset diversity on object detection and semantic segmentation tasks.}
\center
\footnotesize
\begin{tabular*}{1\linewidth}{
p{1.4cm}<{}
p{2.4cm}<{\centering}
p{2.2cm}<{\centering}
p{1.3cm}<{\centering}
}
\toprule
\bf{Task} & \bf{Pre-training Dataset} & \bf{Evaluation Dataset} & \bf{Result} \\\midrule
\multirow{4}{*}{Detection} 
& - & WaterScenes & 59.8 \\
& SMD & WaterScenes & 59.9 (0.1$\uparrow$) \\ \cmidrule(lr){2-4}
& - & SMD & 55.8 \\
& WaterScenes & SMD & 61.5 (\bf{5.7}$\uparrow$) \\ \midrule

\multirow{4}{*}{Segmentation} 
& - & WaterScenes & 85.7 \\
& USVInland & WaterScenes & 86.1 (0.4$\uparrow$) \\ \cmidrule(lr){2-4}
& - & USVInland & 92.5 \\
& WaterScenes & USVInland & 98.3 (\bf{5.8}$\uparrow$) \\
\bottomrule
\end{tabular*}
\vspace{1mm}
\label{tab:domain-gap-1}
\end{table}

To understand the superiority of our new datasets over existing datasets focused on water surfaces, we conduct experiments in two tasks: comparing WaterScenes and SMD \cite{moosbauer2019benchmark} in object detection task and comparing WaterScenes and USVInland \cite{cheng2021we} in semantic segmentation task. 
Specifically, in object detection experiments, we pre-train YOLOv8-N \cite{yolov8} on WaterScenes, followed by training and testing on SMD dataset.
Table \ref{tab:domain-gap-1} shows a remarkable performance improvement of 5.7\% mAP$_{50}$ using our WaterScenes as the pre-training dataset. 
In contrast, there is only 0.1\% increase of mAP$_{50}$ when we use SMD as the pre-training dataset while training and testing on WaterScenes. 
This stark contrast highlights the superior generalization capabilities of a model trained on WaterScenes compared to scenarios in SMD dataset.
Similarly to object detection experiments, we perform SegFormer-B0 \cite{xie2021segformer} on WaterScenes and USVInland, achieving 5.8\% mIoU improvement leveraging WaterScenes as the pre-training dataset.
Experimental results from both object detection and semantic segmentation indicate the diversity compared to existing datasets as well as the inherent value derived from using WaterScenes as a pre-training resource. 

\subsection{Limitations} 
Although WaterScenes represents the first multi-task 4D radar-camera fusion dataset on water surfaces, offering valuable resources to this field, some limitations still exist in our work. 
Given that we aim to explore a low-cost and robust perception approach using radar and camera modalities, we excluded high-definition LiDAR. Thus, object detection is limited to 2D annotations, as sparse radar point clouds cannot replace LiDAR for 3D bounding box annotation. 
Instead, radar data serves as a feature pattern to assist the camera in fusion-based 2D object detection rather than independently completing reliable detection tasks. 
In addition, we mainly focused on providing a foundational baseline for radar-camera fusion on water surfaces using our newly introduced dataset. The accuracy improvement might seem insignificant due to the absence of advanced fusion techniques. Nevertheless, our baseline serves as an essential starting point, and more advanced fusion algorithms could yield significantly higher accuracy levels.


\subsection{Future Works} 
%
As a relatively unexplored field, autonomous driving on water surfaces presents several potential research directions. 
Compared to autonomous driving on road surfaces, perception challenges encountered on water surfaces are more daunting and unpredictable, including water splashes, mirror-like reflections, adverse lighting and weather conditions. With our WaterScenes dataset containing diverse scenarios and environmental conditions, researchers can customize algorithms to address the challenges of camera-based perception on water surfaces.
Additionally, current perception models for autonomous driving emphasize multi-modal fusion and multi-task learning trends \cite{hu2023planning, ye2023fusionad}. A high-generalization, reusable fusion approach can reduce operational costs and power consumption, thus improving the inference speed \cite{liang2022effective}. With our diverse collection of radar and camera data captured from real-world water environments, constructing a multi-task and multi-sensor robust perception model suitable for water surfaces is an interesting and potential research direction. 



\vspace{2.5mm}
\section{Conclusion}
\label{sec:Conclusion}

This work presents a pioneering multi-modal and multi-task dataset that sheds light on previously unexplored 4D radar-camera fusion on water surfaces. Leveraging the complementary advantages of radar and camera sensors, our WaterScenes dataset enables multi-attribute and all-weather perception of the water environment.
We evaluate SOTA algorithms on camera image modality, radar point cloud modality and radar-camera fusion modality on WaterScenes, generating insights into water surface perception that were previously unknown. 
Experimental results demonstrate the value of the dataset for further investigation and also indicate that the 4D radar-camera combination is a robust solution for USVs on water surfaces. 
Without optimization on popular models, radar-camera fusion can actually improve detection performance, especially in adverse lighting and weather conditions.
Overall, the presented WaterScenes offers a valuable resource for researchers interested in autonomous driving on water surfaces and aims to motivate novel ideas and directions for the development of water surface perception algorithms.

\section*{Acknowledgments}

This research was funded by the Suzhou Municipal Key Laboratory for Intelligent Virtual Engineering (SZS2022004), the Suzhou Science and Technology Project (SYG202122), the Research Development Fund of XJTLU (RDF-19-02-23), XJTLU AI University Research Centre, Jiangsu Province Engineering Research Centre of Data Science and Cognitive Computation at XJTLU and SIP AI innovation platform (YZCXPT2022103).
This work received financial support from Jiangsu Industrial Technology Research Institute (JITRI) and Wuxi National Hi-Tech District (WND).

\bibliographystyle{IEEEtran}
\bibliography{citations,others}

\begin{IEEEbiography}[{\includegraphics[width=1in,height=1.25in,clip,keepaspectratio]{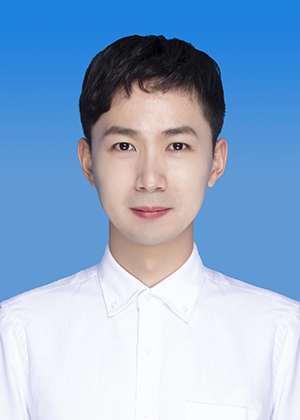}}]{Shanliang Yao} (Student Member, IEEE) received the B.E. degree in 2016 from the School of Computer Science and Technology, Soochow University, Suzhou, China, and the M.S. degree in 2021 from the Faculty of Science and Engineering, University of Liverpool, Liverpool, U.K. He is currently a joint Ph.D. student of University of Liverpool, Xi'an Jiaotong-Liverpool University and Institute of Deep Perception Technology, Jiangsu Industrial Technology Research Institute. His current research is centered on multi-modal perception using deep learning approach for autonomous driving. He is also interested in robotics, intelligent vehicles and intelligent transportation systems. 
\end{IEEEbiography}
\vskip -1cm
\begin{IEEEbiography}
[{\includegraphics[width=1in,height=1.25in,clip,keepaspectratio]{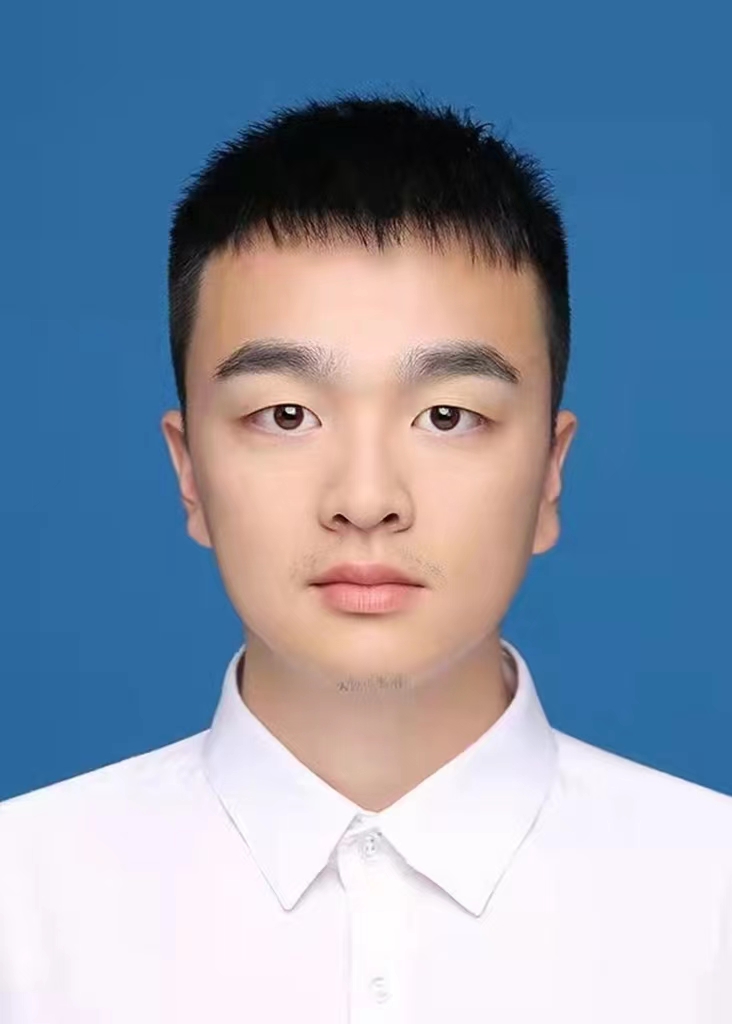}}]{Runwei Guan} (Student Member, IEEE) received his M.S. degree in Data Science from University of Southampton, Southampton, United Kingdom, in 2021. He is currently a joint Ph.D. student of University of Liverpool, Xi'an Jiaotong-Liverpool University and Institute of Deep Perception Technology, Jiangsu Industrial Technology Research Institute. His research interests include visual grounding, panoptic perception based on the fusion of radar and camera, lightweight neural network, multi-task learning and statistical machine learning. He serves as the peer reviewer of IEEE TRANSACTIONS ON NEURAL NETWORKS AND LEARNING SYSTEMS, Engineering Applications of Artificial Intelligence, Journal of Supercomputing, IJCNN, etc.
\end{IEEEbiography}
\vskip -1cm
\begin{IEEEbiography}
[{\includegraphics[width=1in,height=1.25in,clip,keepaspectratio]{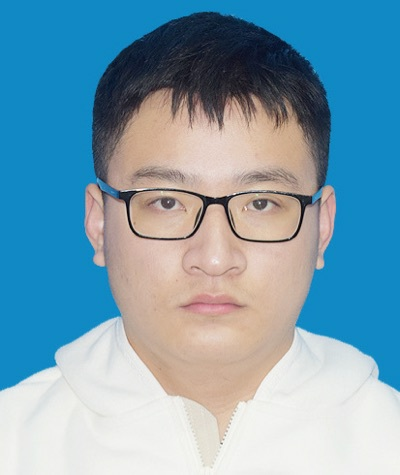}}]{Zhaodong Wu} is an undergraduate student from Xi'an Jiaotong-Liverpool University and will receive his bachelor's degree in 2024. His primary research interests are AI-supported clinical decision support, especially medical image processing, and computer vision for autonomous driving.
\end{IEEEbiography}
\vskip -1cm
\begin{IEEEbiography}
[{\includegraphics[width=1in,height=1.25in,clip,keepaspectratio]{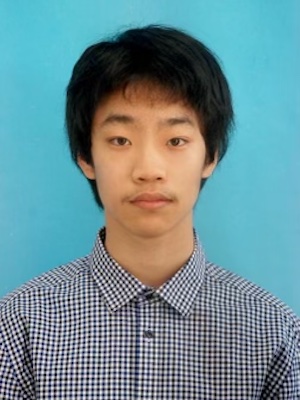}}]{Yi Ni} is an undergraduate student at Xi'an Jiaotong-Liverpool University (XJTLU), majoring in Information and Computing Science. His research interests include computer vision, panoptic perception based on the fusion of radar and camera, medical image segmentation and classification, and multi-task learning. He has participated in research including multi-modal perception based on the fusion of radar and camera, predicting neurological recovery from coma after cardiac arrest using EEG recordings, and cervical spine fracture detection through two-stage approach of mask segmentation and windowing.
\end{IEEEbiography}
\vskip -1cm
\begin{IEEEbiography}
[{\includegraphics[width=1in,height=1.25in,clip,keepaspectratio]{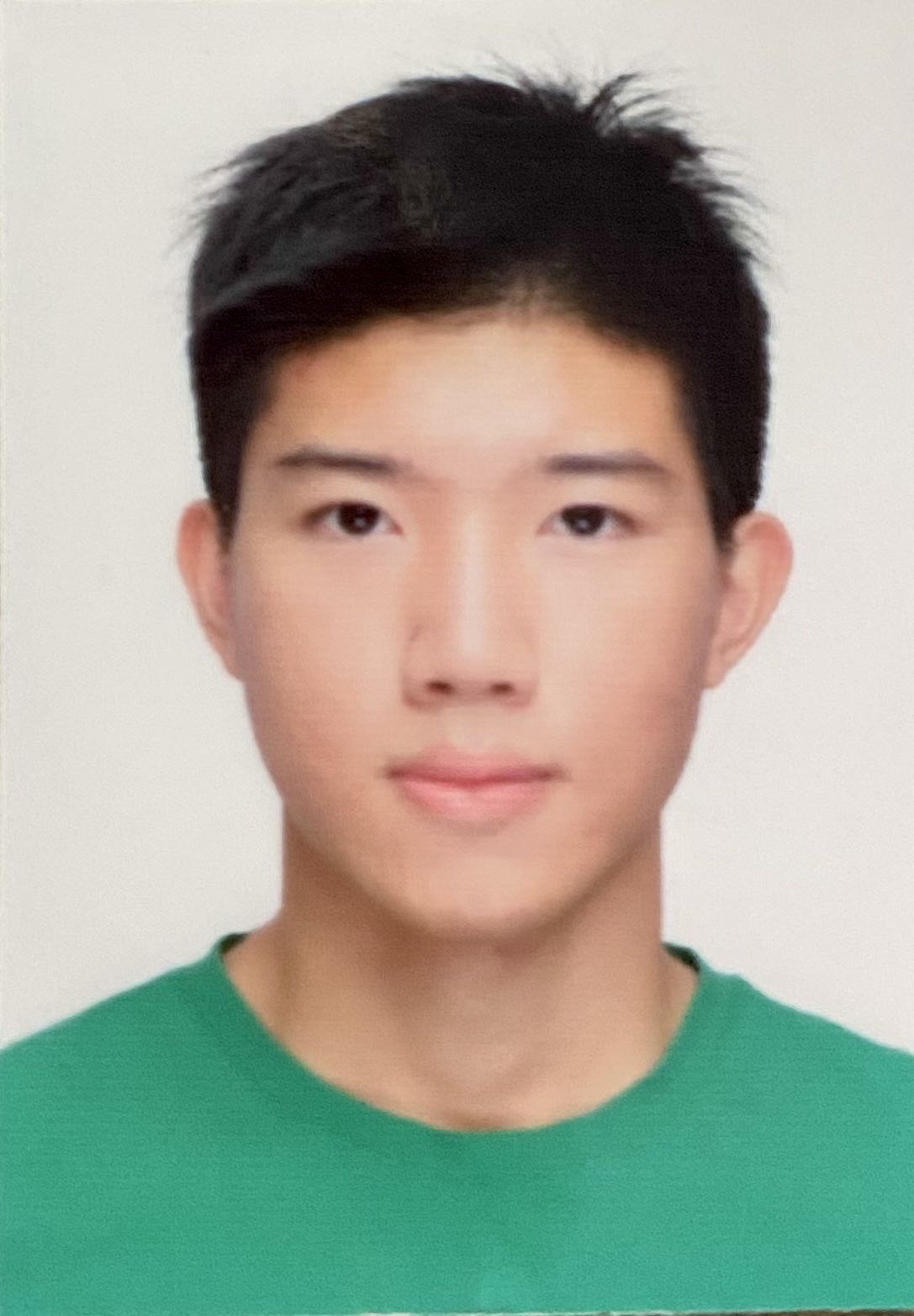}}]{Zile Huang} is currently an undergraduate student at the University of Liverpool, Xi'an Jiaotong-Liverpool University. He is pursuing a degree in Information and Computing Science. His research interests include real-time object detection, multimodal learning, and scene understanding.
\end{IEEEbiography}
\vskip -1cm
\begin{IEEEbiography}[{\includegraphics[width=1in,height=1.25in,clip,keepaspectratio]{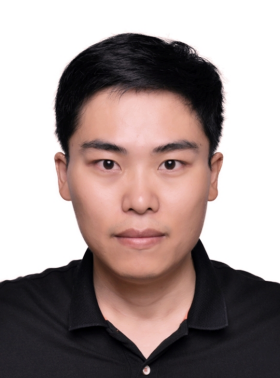}}]{Ryan Wen Liu}
 (Member, IEEE) received the B.Sc. degree (Hons.) in Information and Computing Science from the Department of Mathematics, Wuhan University of Technology, Wuhan, China, in 2009, and the Ph.D. degree from The Chinese University of Hong Kong, Hong Kong, in 2015. He is currently a Professor with the School of Navigation, Wuhan University of Technology. His research interests include intelligent waterborne transportation systems and intelligent marine vehicles. He is an Associate Editor of the \textit{IET Intelligent Transport Systems}, \textit{International Journal of Intelligent Transportation Systems Research}, and \textsc{IEEE Open Journal of Vehicular Technology}.
\end{IEEEbiography}
\vskip -1cm
\begin{IEEEbiography}
[{\includegraphics[width=1in,height=1.25in,clip,keepaspectratio]{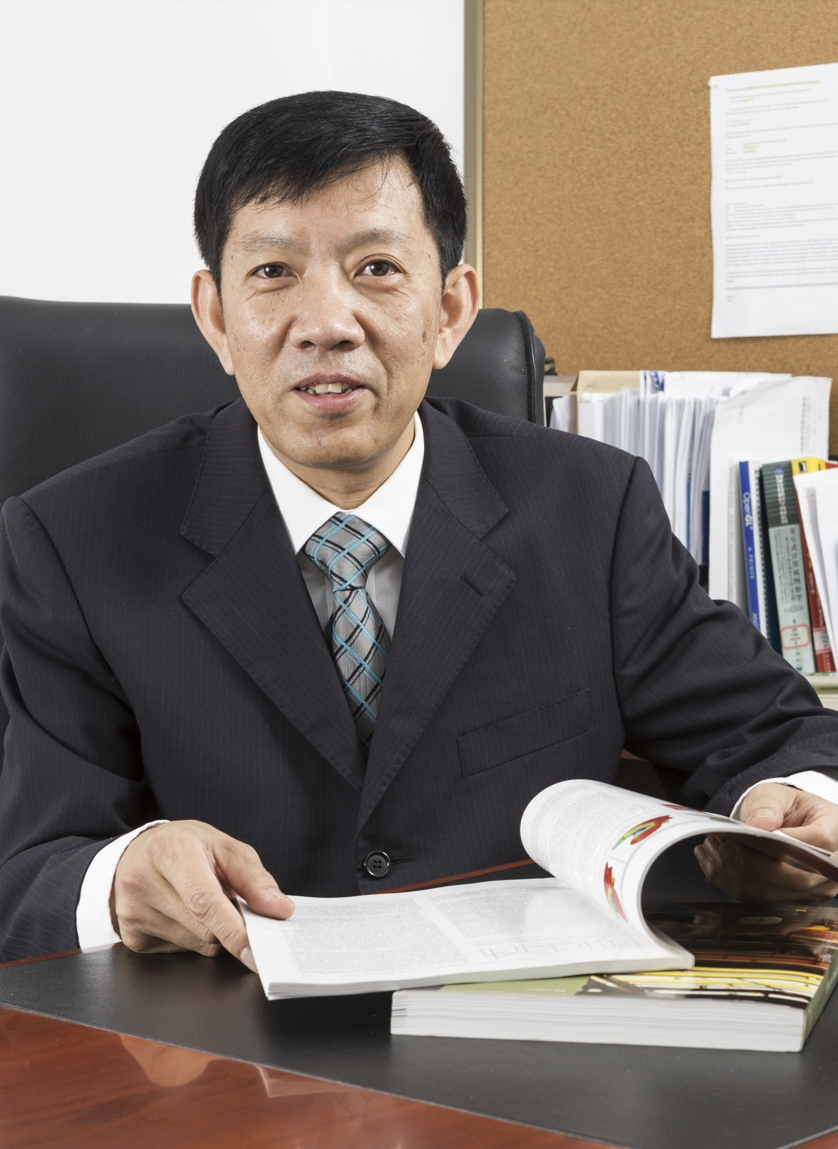}}]{Yong Yue}
Fellow of Institution of Engineering and Technology (FIET), received the B.Eng. degree in mechanical engineering from Northeastern University, Shenyang, China, in 1982, and the Ph.D. degree in computer aided design from Heriot-Watt University, Edinburgh, U.K., in 1994. He worked in the industry for eight years and followed experience in academia with the University of Nottingham, Cardiff University, and the University of Bedfordshire, U.K. He is currently a Professor and Director with the Virtual Engineering Centre, Xi'an Jiaotong-Liverpool University, Suzhou, China. His current research interests include computer graphics, virtual reality, and robot navigation.
\end{IEEEbiography}
\vspace{-1cm}
\begin{IEEEbiography}[{\includegraphics[width=1in,height=1.25in,clip,keepaspectratio]{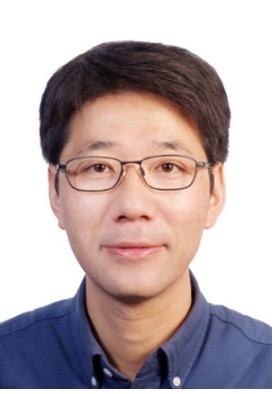}}]{Weiping Ding} (M'16-SM'19) received the Ph.D. degree in Computer Science, Nanjing University of Aeronautics and Astronautics, Nanjing, China, in 2013. In 2016, He was a Visiting Scholar at National University of Singapore, Singapore. From 2017 to 2018, he was a Visiting Professor at University of Technology Sydney, Australia. He is a Full Professor with the School of Information Science and Technology, Nantong University, Nantong, China, and also the supervisor of Ph.D postgraduate by the Faculty of Data Science at City University of Macau, China. His main research directions involve deep neural networks, multimodal machine learning, and medical images analysis. He ranked within the top 2\% Ranking of Scientists in the World by Stanford University (2020-2023). He has published over 250 articles, including over 100 IEEE Transactions papers. His fifteen authored/co-authored papers have been selected as ESI Highly Cited Papers. He serves as an Associate Editor/Editorial Board member of IEEE Transactions on Neural Networks and Learning Systems, IEEE Transactions on Fuzzy Systems, IEEE/CAA Journal of Automatica Sinica, IEEE Transactions on Intelligent Transportation Systems, IEEE Transactions on Intelligent Vehicles, IEEE Transactions on Emerging Topics in Computational Intelligence, IEEE Transactions on Artificial Intelligence, Information Fusion, Information Sciences, Neurocomputing, Applied Soft Computing. He is the Leading Guest Editor of Special Issues in several prestigious journals, including IEEE Transactions on Evolutionary Computation, IEEE Transactions on Fuzzy Systems, and Information Fusion.
\end{IEEEbiography}
\vspace{-1cm}
\begin{IEEEbiography}
[{\includegraphics[width=1in,height=1.25in,clip,keepaspectratio]{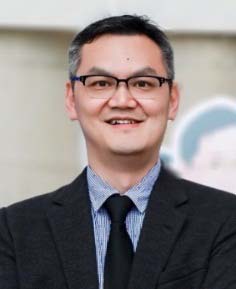}}]{Eng Gee Lim}
(Senior Member, IEEE) received the B.Eng. (Hons.) and Ph.D. degrees in Electrical and Electronic Engineering (EEE) from Northumbria University, Newcastle, U.K., in 1998 and 2002,
respectively. He worked for Andrew Ltd., Coventry, U.K., a leading communications systems company from 2002 to 2007. Since 2007, he has been with Xi'an Jiaotong-Liverpool University, Suzhou, China, where he was the Head of the EEE Department, and the University Dean of research and graduate studies. He is currently the School Dean of Advanced Technology, the Director of the AI University Research Centre, and a Professor with the Department of EEE. He has authored or coauthored over 100 refereed international journals and conference papers. His research interests are artificial intelligence (AI), robotics, AI+ health care, international standard (ISO/IEC) in robotics, antennas, RF/microwave engineering, EM measurements/simulations, energy harvesting, power/energy transfer, smart-grid communication, and wireless communication networks for smart and green cities. He is a Charted Engineer and a fellow of The Institution of Engineering and Technology (IET) and Engineers Australia. He is also a Senior Fellow of Higher Education Academy (HEA).
\end{IEEEbiography}
\vspace{-1cm}
\begin{IEEEbiography}[{\includegraphics[width=1in,height=1.25in,clip,keepaspectratio]{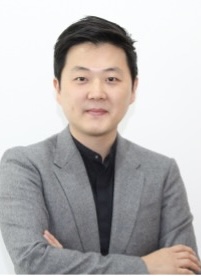}}]{Hyungjoon Seo} (Member, IEEE) received the bachelor's degree in civil engineering from Korea University, Seoul, South Korea, in 2007, and the Ph.D. degree in geotechnical engineering from Korea University in 2013. In 2013, he worked as a research professor in Korea University. He served as a visiting scholar at University of Cambridge, Cambridge, UK, and he worked for engineering department in University of Cambridge as a research associate from 2014 to 2016. In August 2016, he got an assistant professor position in the Department of Civil Engineering at the Xi'an Jiaotong Liverpool University (XJTLU), China. He has been an assistant professor at the University of Liverpool, UK, from 2020. His research interests are monitoring using artificial intelligence and SMART monitoring system for infrastructure, soil-structure interaction (tunneling, slope stability, pile), Antarctic survey and freezing ground. 
Hyungjoon is the director of the CSMI (Centre for SMART Monitoring Infrastructure), CSMI is collaborating with University of Cambridge, University of Oxford, University of Bath, UC Berkeley University, Nanjing University, and Tongji University on SMART monitoring. He presented a keynote speech at the 15th European Conference on Soil Mechanics and Geotechnical Engineering in 2015. He is currently appointed editor of the CivilEng journal and organized two international conferences. He has published more than 50 scientific papers including a book on Geotechnical Engineering and SMART monitoring. 
\end{IEEEbiography}
\vspace{-1cm}
\begin{IEEEbiography}[{\includegraphics[width=1in,height=1.25in,clip,keepaspectratio]{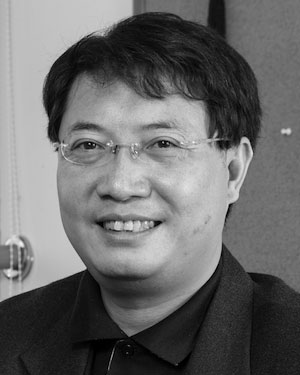}}]{Ka Lok Man}
(Member, IEEE) received the Dr. Eng. degree in electronic engineering from the Politecnico di Torino, Turin, Italy, in 1998, and the Ph.D. degree in computer science from Technische Universiteit Eindhoven, Eindhoven, The Netherlands, in 2006. He is currently a Professor in Computer Science and Software Engineering with Xi'an Jiaotong-Liverpool University, Suzhou, China. His research interests include formal methods and process algebras, embedded system design and testing, and photovoltaics.
\end{IEEEbiography}
\vspace{-1cm}
\begin{IEEEbiography}[{\includegraphics[width=1in,height=1.25in,clip,keepaspectratio]{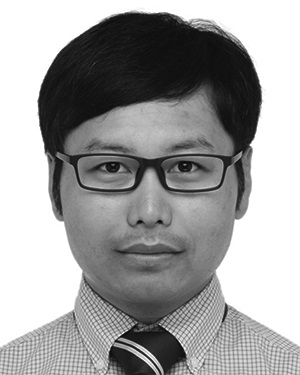}}]{Jieming Ma} received the M.Sc. degree in advanced microelectronic systems engineering from the University of Bristol, UK, in 2010, and received the Ph.D. degree in computer science from the University of Liverpool, UK, in 2014. He is currently working as an Associate Professor at the Xi'an Jiaotong-Liverpool University, China. His research interests include intelligent optimization, machine learning and applications in renewable energy systems.
\end{IEEEbiography}
\vspace{-1cm}
\begin{IEEEbiography}
[{\includegraphics[width=1in,height=1.25in,clip,keepaspectratio]{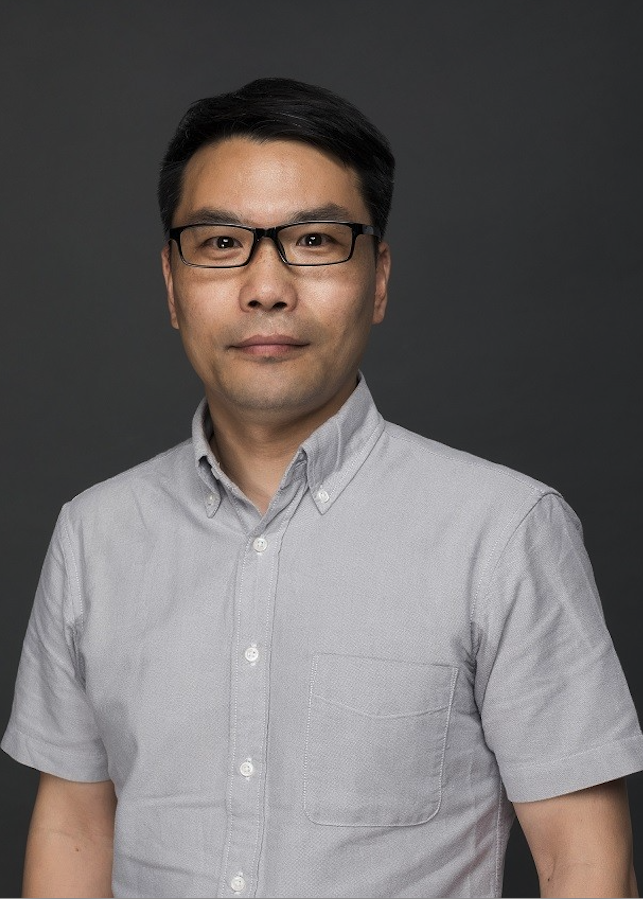}}]{Xiaohui Zhu}
(Member, IEEE) received his Ph.D. from the University of Liverpool, UK in 2019. He is currently an associate professor, Ph.D. supervisor and Programme Director with the Department of Computing, School of Advanced Technology, Xi'an Jiaotong-Liverpool University. He focuses on advanced techniques related to autonomous driving, including sensor-fusion perception, fast path planning, autonomous navigation and multi-vehicle collaborative scheduling. 
\end{IEEEbiography}
\vspace{-1cm}
\begin{IEEEbiography}[{\includegraphics[width=1in,height=1.25in,clip,keepaspectratio]{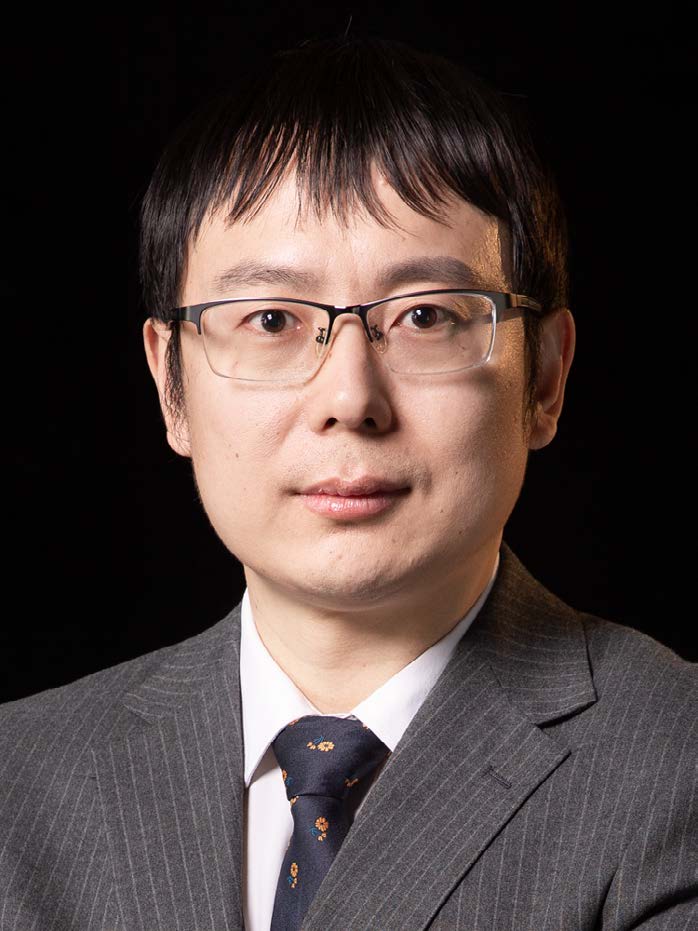}}]{Yutao Yue}
(Member, IEEE) is an associate professor at the Artificial Intelligence Thrust and Intelligent Transportation Thrust of Hong Kong University of Science and Technology (Guangzhou). He received his Bachelor’s degree from the University of Science and Technology of China, and Master and PhD degree from Purdue University. He has a dual background in academia and industry, as the team leader of Guangdong Province Introduced Innovation Scientific Research Team, senior scientist of Kuang-Chi Group, and the founder of the Institute of Deep Perception Technology of JITRI. His research interests include multimodal perception fusion, machine consciousness, artificial general intelligence, causal emergence, etc. He has been engaged in scientific research and technology industrialization for over 20 years. He has co-invented 354 granted Chinese patents, 18 USA patents, and 7 EU patents. He has led 6 major research projects with a total funding of nearly 130 million RMB. He has published over 60 papers, advised 13 postdoc research fellows, and received multiple awards including Wu Wenjun Artificial Intelligence Science and Technology Award.
\end{IEEEbiography}

%
%
%
%
%
%
%
%
%
%
%
%
%
%

\end{document}